\documentclass[lettersize,journal]{IEEEtran}
\usepackage{amsmath,amsfonts}
\usepackage{algorithmic}
\usepackage{algorithm}
\usepackage{array}
\usepackage[caption=false,font=footnotesize,labelfont=sf,textfont=sf]{subfig}
\usepackage{textcomp}
\usepackage{stfloats}
\usepackage{url}
\usepackage{verbatim}
\usepackage{graphicx}
\usepackage{cite}
\usepackage{comment}
\usepackage{xcolor}
\usepackage{soul}
\begin{document}

\title{Federated Transfer-Ordered-Personalized Learning for Driver Monitoring Application}

\author{Liangqi Yuan,~\IEEEmembership{Student Member,~IEEE},
        Lu Su,~\IEEEmembership{Member,~IEEE},
        Ziran Wang,~\IEEEmembership{Member,~IEEE}
\thanks{Manuscript received January 11, 2023.}
\thanks{L. Yuan, L. Su, and Z. Wang are with the College of Engineering, Purdue University, West Lafayette, IN 47907, USA (e-mail: liangqiy@purdue.edu; lusu@purdue.edu; ryanwang11@hotmail.com).}}

\markboth{IEEE Internet of Things Journal}%
{Shell \MakeLowercase{\textit{et al.}}: A Sample Article Using IEEEtran.cls for IEEE Journals}


\maketitle

\begin{abstract}
Federated learning (FL) shines through in the internet of things (IoT) with its ability to realize collaborative learning and improve learning efficiency by sharing client model parameters trained on local data. Although FL has been successfully applied to various domains, including driver monitoring applications (DMAs) on the internet of vehicles (IoV), its usages still face some open issues, such as data and system heterogeneity, large-scale parallelism communication resources, malicious attacks, and data poisoning. This paper proposes a federated transfer-ordered-personalized learning (FedTOP) framework to address the above problems and test on two real-world datasets with and without system heterogeneity. The performance of the three extensions, transfer, ordered, and personalized, is compared by an ablation study and achieves 92.32$\%$ and 95.96$\%$ accuracy on the test clients of two datasets, respectively. Compared to the baseline, there is a 462$\%$ improvement in accuracy and a 37.46$\%$ reduction in communication resource consumption. The results demonstrate that the proposed FedTOP can be used as a highly accurate, streamlined, privacy-preserving, cybersecurity-oriented, and personalized framework for DMA.
\end{abstract}

\begin{IEEEkeywords}
Federated learning, internet of things (IoT), driver monitoring, privacy protection, personalization.
\end{IEEEkeywords}

\section{Introduction}
\IEEEPARstart{W}{ith} the rapid development of sensing, computing, and communication technologies, the internet of things (IoT) is a popular solution to solve problems in industry, agriculture, energy, transportation, etc. However, privacy issues in IoT are often a significant concern have been raised due to the intrusive behavior of sensors \cite{yang2017survey}. Specifically for the internet of vehicles (IoV), it massively parallels each vehicle and various sensors it carries, including global positioning system (GPS), radar, camera, light detection and ranging (LiDAR), etc., enabling pedestrian detection \cite{cao2021handcrafted}, automated driving \cite{kuutti2018survey}, mobility digital twins \cite{wang2022mobility}, and other transportation applications. Federated learning (FL) has received extensive attention for protecting user privacy by sharing only model weights and not including users' raw data. FL is widely known for its successful business case in Google mobile keyboard prediction \cite{hard2018federated}. Nowadays, it has also become one of the mainstream and thriving solutions for privacy protection and efficient learning.

\subsection{Federated Learning and Related Work}
\label{Sec. Federated Learning and Related Work}
FL is a potentially feasible solution to the privacy problem in IoT, which is able to avoid the proliferation, distribution, and exchange of local client data by sharing model parameters after training the model on local client data. FL frameworks are widely used in mobile service \cite{zhou20222d}, healthcare \cite{dayan2021federated,rieke2020future}, industrial \cite{hao2019efficient,lu2019blockchain}, IoV \cite{du2020federated,kong2021federated}, etc., due to their usages of large scale and personalized data in an efficient and privacy-preserving way. Although FL has significant contributions to massively parallel devices and computations, it still has a notable drawback in that it cannot efficiently handle non-independent and identically distributed (non-i.i.d.) data. It is required to customize the applicable FL framework according to the features, resources, and constraints possessed by users, data, clients, and servers.

Non-i.i.d. data and heterogeneity have always been a challenge, and a key to research in FL \cite{sattler2019robust,karimireddy2020scaffold,horvath2021fjord}. Non-i.i.d. data is a common phenomenon for real-world clients that are scattered and not interoperable: Taking IoV as an example, each driver is heterogeneous as a client. FedAvg \cite{mcmahan2017communication}, as one of the first proposed feasibility methods, has been the subject and center of research. FedAvg averages all local models to get the global model so that the local model may deviate far from the global optimum in the parameter space leading to some limitations in FedAvg. It is necessary to ensure that the local model does not deviate from the global model (prevent overfitting) and, simultaneously, that the local model can effectively learn the local client dataset (prevent underfitting). Based on FedAvg, FedProx \cite{li2020federatedFedProx} is proposed to limit the deviation of the local model from the global model by adding a proximal term. 

Besides considering accuracy, the FL framework in IoT should not underestimate communication and training resource constraints, cybersecurity, and ubiquity. Some of the recent surveys summarized challenges, threats, and solutions of the FL decentralization paradigm for IoT, including limited computing power, unreliable and limited availability, local training, accuracy, communication overhead, etc. \cite{ghimire2022recent,li2020federatedchallenge,niknam2020federated,lyu2020threats,kairouz2021advances,li2021survey}.

Transfer and edge learning are popular solutions to reduce communication resource consumption in FL frameworks. Zhang \textit{et al.} \cite{zhang2022privacy} performed a federated transfer learning framework to detect driver drowsiness, where transfer learning was employed to save the communication cost in the FL framework. Su \textit{et al.} \cite{su2021secure} introduced edge servers as a collaborative mechanism, where local models were aggregated in the edge server and then sent to the global server to aggregate the global model. The benefit of the additional edge server was that the communication between massively parallel clients and the edge server was consumed because the edge server was geographically close to the clients. High latency and intermittent connections could be mitigated. In addition, the edge server could also provide personalized aggregated local models due to the similarity of geographically adjacent clients. 

Cyber attack is a problem that cannot be ignored for FL frameworks. Sun \textit{et al.} \cite{sun2021data} developed an attack method for FL framework in IoT, in which a bi-level optimization framework was proposed to compute optimal poisoning attacked FL framework, including direct, indirect, and hybrid attacks. Meanwhile, Zhang \textit{et al.} \cite{zhang2020poisongan} utilized a generative adversarial network (GAN)-based approach to attack the FL framework, especially since the attacker did not need any prior knowledge to carry out the attack. 

Personalization is a common approach for FL frameworks to improve applicability for diverse users \cite{tan2022towards}. Fallah \textit{et al.} \cite{fallah2020personalized} proposed a personalized variant of the FL, which allowed clients to perform several gradient descent iterations on an initial global model using local data to obtain a personalized local model. Wu \textit{et al.} \cite{wu2020fedhome} explored a cloud edge-based personalized FL framework for in-home health monitoring, which addressed the problem that a single global model performed poorly on a specific client. Since the global model could only capture the common features of all clients, it lacked the ability to analyze fine-grained information of specific clients. Ma \textit{et al.} \cite{ma2022layer} proposed a personalized FL framework of layer-wised weighted aggregation, which determines mutual contribution factors through a hypernetwork on the server to identify similarity between users at the layer granularity.

\subsection{Federated Learning in Driver Monitoring Applications} 
\label{Sec. Federated Learning in Driver Monitoring Applications}
Driver monitoring application (DMA) in IoV is adopted as the research direction in this paper due to its real and visual image data, valuable application scenarios, and relatively blank research area. DMA also has challenges in terms of driver privacy issues, communication, and diverse and personalized driver state and behavior. Related DMA literature covers a wide variety of devices with algorithms to achieve different purposes, such as dangerous state detection \cite{kashevnik2019methodology}, driver emotion recognition \cite{zepf2020driver}, driver lane change inference \cite{xing2019driver}, etc. Compared to other methods \cite{masood2020security,kuutti2020survey,ramzan2019survey}, FL not only highlights efficient learning but also effectively protects the privacy of driver, passenger, and pedestrian biometric information, driving routes, and confidential driving areas such as military installations. 

In this paper, we introduce and adapt FL to DMA. Although some FL frameworks exist for DMA, they all suffer from some critical problems. Doshi \textit{et al.} \cite{doshi2022federated} proposed a FL edge-device framework to obtain a global model by aggregation feature representations and obtained considerable accuracy in recognizing driver activities. For the i.i.d. setting, the dataset was partitioned for each edge node in a random way, while for the non-i.i.d. setting, the dataset was assigned selectively. Zhao \textit{et al.} \cite{zhao2023fedsup} proposed a FL framework to monitor fatigue driving, where the non-i.i.d. setting was simulated by controlling the number of images per client. The above FL frameworks for DMA did not really take into account the actual situation of the application but artificially created a simulation scenario. Therefore, there is an urgent need for realistic analysis and research for real-world DMA, considering that the user (driver) should exist independently and be non-interoperable with different clients (vehicles). Moreover, in addition to the necessity of test datasets, the test client is also a critical evaluation criterion, which can reflect the universality of the FL framework. We summarize the existing neglects and challenges in the current FL for DMA frameworks as follows.
\begin{itemize}
\item Clients in FL for DMA frameworks are often defined in unreasonable and incomprehensible forms. A real and natural definition of a client should be a driver or a vehicle.
\item There is no paper proposing to test on a testing client (not involved in the training process), which lacks universal testing for the FL framework.
\item For the DMA scenario, there is a great diversity and individuality of driver behaviors, postures, and facial expressions, which call for more personalized studies than other general IoV scenarios.
\item Similarly, DMA also has diverse scenarios, including diverse vehicle models, interior colors, seat positions, etc., which will greatly increase the learning difficulty.
\end{itemize}

\subsection{Proposed Solution and Contribution} 
\label{Sec. Proposed Solution and Contribution}
\begin{figure}[t]
\centering
\centerline{\includegraphics[width=\linewidth]{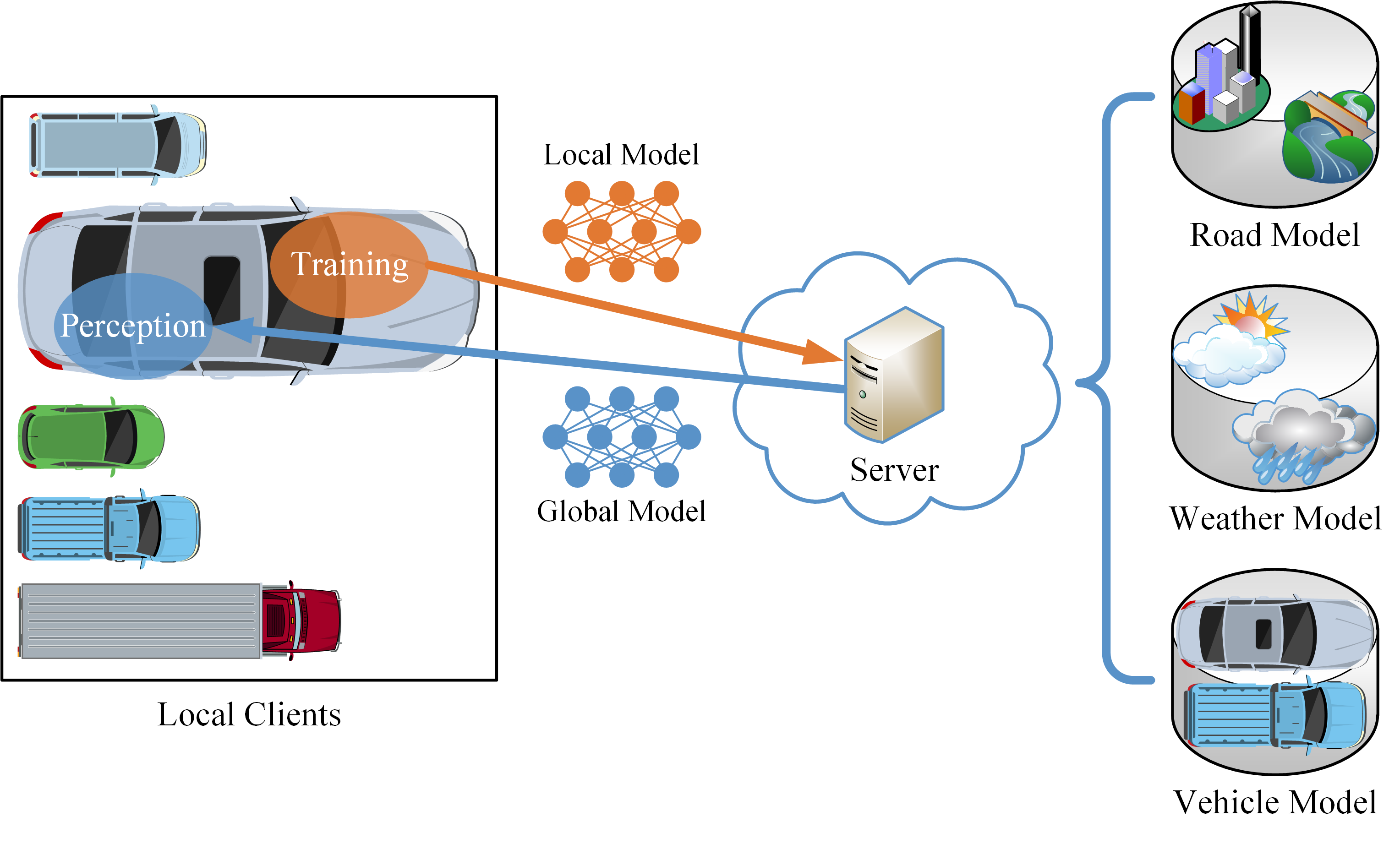}}
\caption{Structure illustration of a FL framework for IoV. The server interacts with the local client and saves different scenarios as different models. Transparent neurons are non-trainable parameters, and non-transparent neurons are trainable parameters.}
\label{Fig. FedTOP structure}
\end{figure}

In this paper, we aim to propose a FL framework applicable and specific to practical applications in IoV, especially DMA, where an imaginary FL framework for IoV is illustrated in Fig. \ref{Fig. FedTOP structure}. Each local client, i.e., vehicle, includes a training module and a perception module. The training module uploads the model parameters to the server after learning and training the local data. After aggregation and optimizing the parameters of the local client models, the server downloads the global model parameters to the perception module in the local client. Moreover, transfer learning can be used to reduce the number of trainable parameters, resulting in reduced communication consumption. The server can save different global models for different scenarios, such as road types, weather types, and vehicle types, so that the model can have better applicability.

Therefore, a federated transfer-ordered-personalized learning (FedTOP) framework is proposed to address the problems of accuracy, cybersecurity, communication resources, and diversified scenarios. In addition to the transfer-extension shown in Fig. \ref{Fig. FedTOP structure}, the FedTOP framework also enhances robustness and cybersecurity by orderly dropout clients due to their possible overfitting and poisoning of the data. Furthermore, the FedTOP framework is able to remarkably improve accuracy by adapting all clients through personalized-extension. The contributions of this paper are:
\begin{itemize}
\item For realistic problems and usage scenarios in DMA, we propose a feasible FL framework FedTOP, realizing privacy protection, high accuracy, low communication requirements, cybersecurity, and pervasiveness. To the best of our knowledge, this is one of the first papers to establish a feasible FL framework for DMA. 
\item The proposed FedTOP framework is tested on two real-world driver monitoring datasets with and without system heterogeneity, systematically characterizing system heterogeneity in real-world datasets and achieving considerable accuracies with 92.32$\%$ and 95.96$\%$, respectively.
\item The experiments highlight a realistic and natural client setup, i.e., drivers and vehicles are naturally formed as clients. Moreover, we innovatively propose evaluation criteria for training and testing clients to test the generalization ability of the proposed FedTOP on different clients.
\item Through an ablation study, we demonstrate the performance and utility of the transfer, ordered, and personalized extensions. These detachable extensions can be selectively installed according to the task description, and the FL framework combined with different extensions can effectively adapt to different IoT application scenarios.
\end{itemize}

The presentation of this paper is as follows. The problem statement and proposed solution are described in Section \ref{Sec. Methodologies}. The experimental setup, heterogeneity, and results have been demonstrated in Section \ref{Sec. Experiment and Results}. Section \ref{Sec. Discussion} discusses the performances of three extensions of the proposed framework, followed by Section \ref{Sec. Conclusion} summarizing the paper and expounding on future work.

\section{Methodologies}
\label{Sec. Methodologies}

\subsection{Problem Statement}
\label{Sec. Problem Statement}
FL frameworks are able to protect privacy, increase training efficiency, and save communication resources by sharing only model parameters in IoT. In this paper, the FL framework is used to solve a driver activity classification task in DMA. Clients in real-world IoT are independent and heterogeneous due to the presence of only a minimal number of users per client. Considering the more general application scenarios, the global model $\omega$ for training clients $C$ aggregation needs to be compatible with non-training clients $C'$ in addition to $C$. The data of each client $D_c$ is non-i.i.d. when the data is not interoperable. We can consider a nested model,
\begin{equation}
    L_c = \omega_c(D_c),
\label{Eq. nested model}
\end{equation}
where $\omega_c$ is the classifier model corresponding to client $c \in C$. $D_c \in \mathbb{R}^{n_c \times i \times j \times d}$ is the image set with $n_c$ samples, $i$ rows, $j$ columns, and $d$ channels. $L_c \in \mathbb{Z}^{n_c}$ is the corresponding label set. The global model $\omega$ is obtained by aggregating, e.g., averaging the weights of the local models,
\begin{equation}
    \omega = \sum_{c \in C} p_c\omega_c = \mathbb{E}[\omega_c | c \in C],
\label{Eq. aggregating}
\end{equation}
where $p_c \in [0,1]$ is a weight density function of clients, for which $\sum p_c=1$, $p_c$ will be assigned according to the number of samples. Therefore, the optimization problem of the FL algorithm can be formulated as minimizing the global loss, which is equivalent to minimizing the sum of the local losses,
\begin{equation}
    \min_\omega \mathcal{L}(\omega) = \sum_{c \in C}p_c\mathcal{L}(\omega_c) = \mathbb{E}[\mathcal{L}(\omega_c) | c \in C],
\label{Eq. global loss}
\end{equation}
where $\mathcal{L}$ is the loss function that will be assigned. 

\begin{figure}[t]
\centering
\centerline{\includegraphics[width=\linewidth]{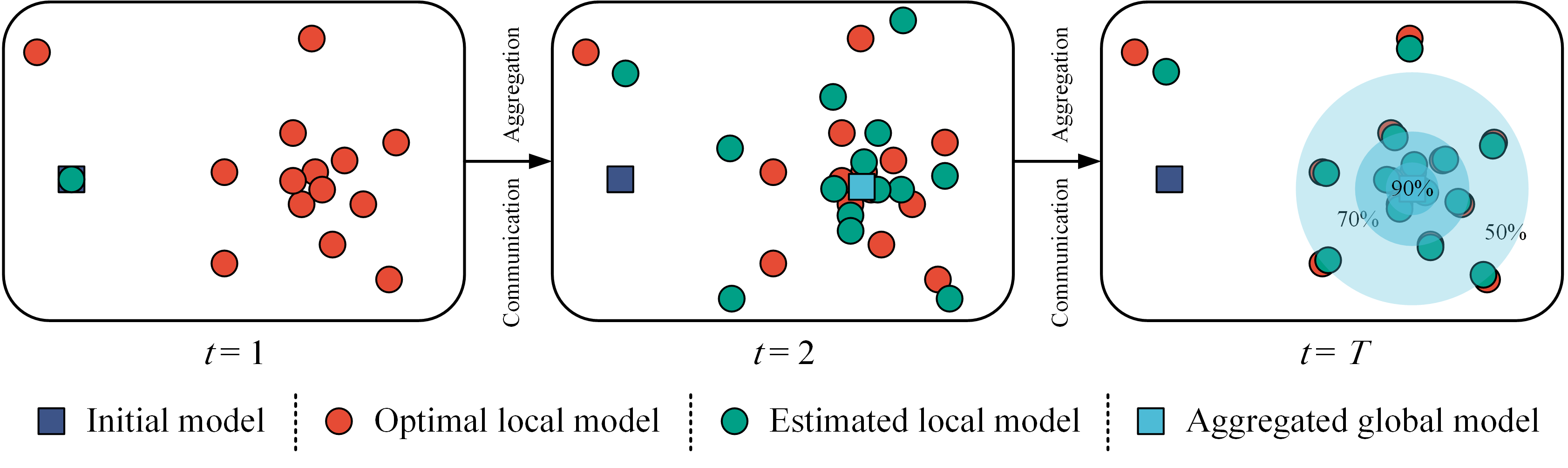}}
\caption{Illustration of the FL algorithm finds the optimal global model solution in the parameter space. The shaded areas are accuracy contour areas. The farther the optimal local model dissociates from the global model, the lower the client accuracy. Local models enclosed by shaded areas have similar accuracies.}
\label{Fig. parameter space}
\end{figure}

For real-world classification tasks, we assume that the distribution of the local model in the parameter space presents a multivariate Normal distribution $\omega_c \sim \mathcal{N}\left(\mu_\omega, \sigma^2_\omega \right)$, where $\mu_\omega$ is mean of all local models, and $\sigma^2_\omega$ is the variance of all local models. Fig. \ref{Fig. parameter space} shows the process of the FL algorithm finding the optimal solution of the global model in the parameter space. After the initial model is trained locally, communicated, and aggregated globally, the final global model will be obtained by averaging and can be estimated as $\hat{\omega} = \mu_\omega$. Especially in the large-scale parallel application scenarios of IoT, according to the law of large numbers, $\hat{\omega} = \mu_\omega = \omega^\ast$ is an unbiased estimation. 

However, there are still some defects in the method of obtaining the global model through average aggregation. Firstly, we can confirm that there is enormous system heterogeneity in IoT, and the global model cannot ensure high accuracy for all clients. Secondly, we inevitably need a measure to prevent system heterogeneity and potential attacks and poisoning. As shown in Fig. \ref{Fig. parameter space}, the farther the optimal local model is from the global model, the lower the accuracy, and vice versa. Therefore, it is conceivable that in the FL problem with heterogeneity, the clients' accuracy will also obey a Normal distribution.

\subsection{Proposed Solution}
\label{Sec. Proposed Solution}

According to the problem statement, we propose a FedTOP algorithm to address all of the following issues. First, the aggregation of global models needs to be more stable, which can be achieved by preventing the overfitting of local models. Second, considering the actual communication situation in IoT, we propose a transfer learning method to reduce the trainable parameters and hence reduce communication requirements. Third, the global model should have the ability to resist interference, attacks, and data poisoning, which can be achieved by orderly dropping out local models with a large loss. Fourth, a global model cannot take into account the situation of all clients, especially in the presence of data and system heterogeneity. Therefore, we recommend personalizing the global model to suit all the training and testing clients.

We refer to FedProx \cite{li2020federatedFedProx} using a proximal term to prevent local models $\omega_c$ from deviating from the global model $\omega$. In which the proximal item $\mathcal{L}_p$ that computes the distance between the local and global model is added to the loss function,
\begin{equation}
    \mathcal{L}_p= \frac{\mu}{2}\|\omega_c-\omega\|^2,
\label{Eq. loss proximal term}
\end{equation}
where $\mu$ is deviation coefficient, $\omega_c$ is local client model parameters, and $\omega$ is global model parameters. The overall loss function can be updated as
\begin{equation}
    \mathcal{L} = \mathcal{L}_l + \mathcal{L}_p,
\label{Eq. loss function}
\end{equation}
where $\mathcal{L}_l$ is the loss between the true labels and the predicted labels, such as the negative log-likelihood loss used in our experiments. 

\begin{figure}[t]
\centering
\centerline{\includegraphics[width=\linewidth]{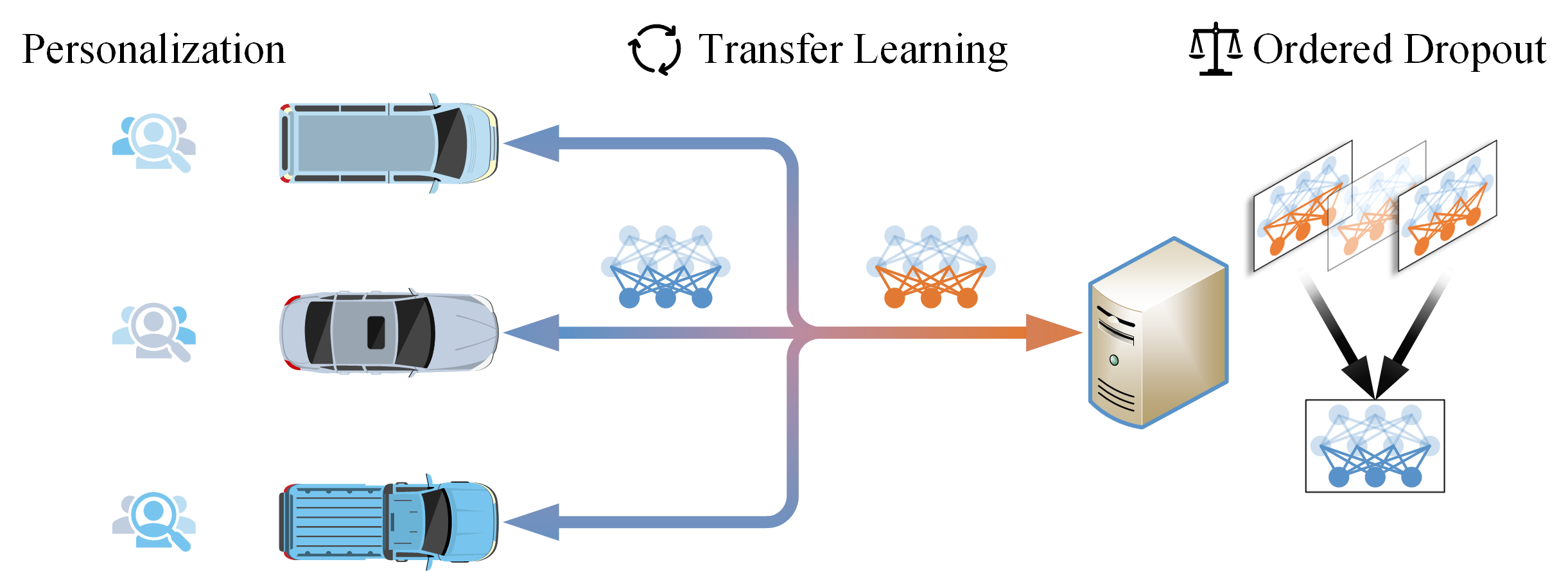}}
\caption{Schematic diagram of the proposed FedTOP system, where transfer, ordered, and personalized extensions are deployed on the communication, server, and client sides, respectively. They operate independently and do not interfere with each other.}
\label{Fig. System Diagram}
\end{figure}

\textit{Transfer-extension} is a common and popular solution in many learning frameworks. In particular, FL frameworks are favored because they can effectively reduce local client training resources and communication resources. In our experiments, the base model is ResNet34 \cite{he2016deep} pre-trained on ImageNet, where only the last residual block and fully connected layer are trainable parameters. Although ImageNet is a large object classification dataset far from DMA images, the lower layers are similar to convolutional neural networks (CNN) and are used to extract image features. Therefore, the upper layers that are used to obtain high-level features and representations are given more attention. The ratio of reduced communication resource requirement in the network is approximately equal to the ratio of non-trainable parameters to total parameters,
\begin{equation}
    \text{Commun}_\downarrow \approx \frac{|\omega_{\text{non-trainable}}|}{|\omega|} = 37.46\%,
\label{Eq. communication resources}
\end{equation}
where $\text{Commun}_\downarrow$ is the reduced communication resource requirement, $|\omega_{\text{non-trainable}}|$ is the number of non-trainable model parameters, and $|\omega|$ is the total number of the model parameters. Therefore, the transfer-extension reduces the communication requirement by 37.46$\%$ by decreasing the trainable parameters.

\textit{Ordered-extension} is for orderly dropout clients with enormous variance, which may be subject to malicious attacks and poisoning, extensive data and system heterogeneity, and model underfitting. These local clients with large losses should be discarded to enhance the generalizability of the global model. Ordered-extension not only enhances accuracy and robustness but also secures the global model. After all of the clients upload the local model parameters and the final training loss to the server, the server only aggregates the $q \in \mathbb{N} \leq |C|$ local models with the lowest loss as the global model. The set of $q$ local models can be expressed as
\begin{equation}
C_q \in q-\arg\min_{c \in C} \mathcal{L}(\omega_c).
\label{Eq. ordered-extension}
\end{equation}
Therefore, only the models of clients in $C_q$ with lower loss will be used to aggregate the global model.

\begin{algorithm}[t]
\small
   \caption{\small{FedTOP}}
   \label{Alg FedTOP}
\begin{algorithmic}
   \STATE {\bfseries Input:} Communication rounds ($T$), training client set ($C$), training epoch ($E$), initial global model ($\omega^1$), loss function ($\mathcal{L}_l$), deviation coefficient ($\mu$), number of ordered clients ($q$)
   \STATE {\bfseries Output:} Trained global model ($\omega^{T}$)
   \FOR{$t=1$ {\bfseries to} $T-1$}
        \FOR{$c \in C$ {\bfseries in parallel}}
            \FOR{$e=1$ {\bfseries to} $E-1$}
                \STATE Backpropagate the loss function and update the local model $\omega_c^{t^{e+1}} \gets \arg\min_{\omega_c^{t^{e}}} \mathcal{L}_l(\omega_c^{t^{e}}) + \frac{\mu}{2}\|\omega_c^{t^{e}}-\omega^t\|^2 $.
            \ENDFOR
            \STATE Update the local model $\omega_c^{t} \gets \omega_c^{t^{E}}$.
            \STATE Client sends $\omega^{t}_c$ to the server.
        \ENDFOR
    \STATE Find a set $C^t_q$ of top-$q$ clients in $C^t$ in term of
loss values: $C^t \in q-\arg\min_{c \in C^t} \mathcal{L}(\omega^{t}_c)$.
    \STATE Server aggregates the $\omega$ as $\omega^{t+1} \gets \frac{1}{q}\sum_{q \in C^t_q} \omega^{t}_q$.
   \ENDFOR
   \STATE Send $\omega^{T}$ to clients $c \in \{C, C'\}$ do personalization.
\end{algorithmic}
\end{algorithm}

\begin{algorithm}[t]
\small
   \caption{\small{Personalized-extension}}
   \label{Alg Personalization}
\begin{algorithmic}
    \STATE {\bfseries Input:} Training client set ($C$), testing client set ($C'$), personalization epoch ($E$), Trained global model ($\omega^{T}$), loss function ($\mathcal{L}_l$)
    \STATE {\bfseries Output:} Personalized local model ($\omega_c$)
    \FOR{$c \in \{C, C'\}$}
        \FOR{$e=1$ {\bfseries to} $E-1$}
            \STATE Backpropagate the loss function and update the local model $\omega_c^{T^{e+1}} \gets \arg\min_{\omega_c^{T^{e}}} \mathcal{L}_l(\omega_c^{T^{e}})$.
        \ENDFOR
        \STATE Update the personalized local model $\omega_c \gets \omega_c^{T^{E}}$.
    \ENDFOR
\end{algorithmic}
\end{algorithm}

\textit{Personalized-extension} is to promote, popularize, and adapt the global model to the heterogeneity of all clients. As shown in Fig. \ref{Fig. parameter space}, the global model cannot be applied to all clients due to the ubiquitous heterogeneity. The region of interest (ROI) of the model may vary depending on system heterogeneity, such as different camera angles, seat positions, and vehicle structures, resulting in differences in the relative position of the driver in the image. However, personalized-extension proposes to train the global model several times in each client to obtain a more personalized local model to improve accuracy. On the one hand, compared with the traditional FL algorithm, the personalized-extension can significantly and effectively improve accuracy and confidence. On the other hand, compared to the method that only trains locally, the personalized FL algorithm improves the training efficiency and avoids the overfitting of the local model. In particular, the personalized FL algorithm can help and generalize to other non-training clients $C'$, which may have minimal training resources. After receiving the global model, the non-training clients $C'$ can obtain a highly accurate and reliable local model with minimal training. The system diagram of the proposed FedTOP is shown in Fig. \ref{Fig. System Diagram}.

For the proposed FedTOP framework, the client communicates with the server $T$ rounds, and all clients $C$ train $E$ epochs in parallel between each communication. For our preliminary experiments, we set $T=10$ and $E=5$. For transfer-extension, the local model is the transfer learning model of ResNet34 pre-trained on ImageNet. Only the last residual block and fully connected layer are set as trainable parameters. In addition, we add an additional fully connected layer to match the number of our classification categories. Based on FedProx, the activation function of the last layer is LogSoftmax, and the setting of the loss function $\mathcal{L}_l$ is a negative log-likelihood loss. $\omega^1$ is the initial model parameter. The proposed FedTOP is described in Algorithm \ref{Alg FedTOP}, and the personalization process is described in Algorithm \ref{Alg Personalization}.

\begin{figure*}[t]
\centering
\subfloat[SFDDD texting - right 1]{\includegraphics[width=0.25\linewidth]{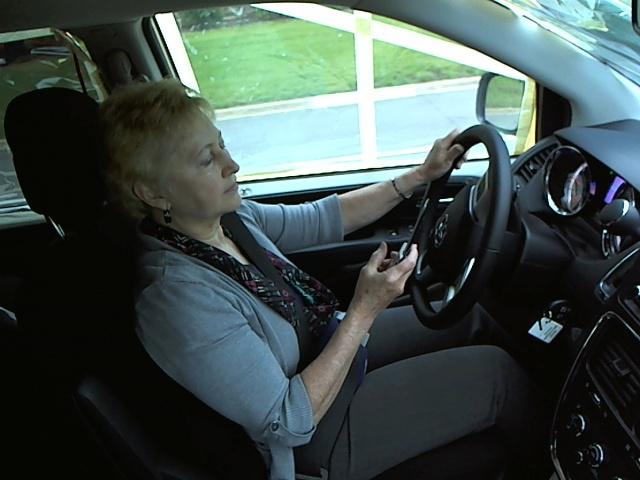}%
\label{Fig. SFDDD 1}}
\hfill
\subfloat[SFDDD texting - right 2]{\includegraphics[width=0.25\linewidth]{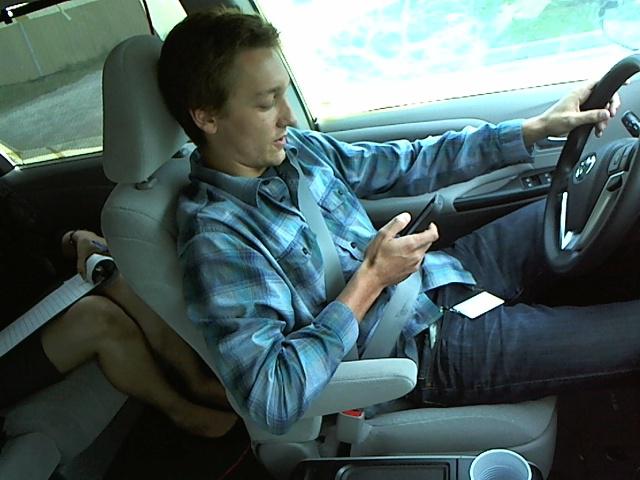}%
\label{Fig. SFDDD 2}}
\subfloat[SFDDD texting - right 3]{\includegraphics[width=0.25\linewidth]{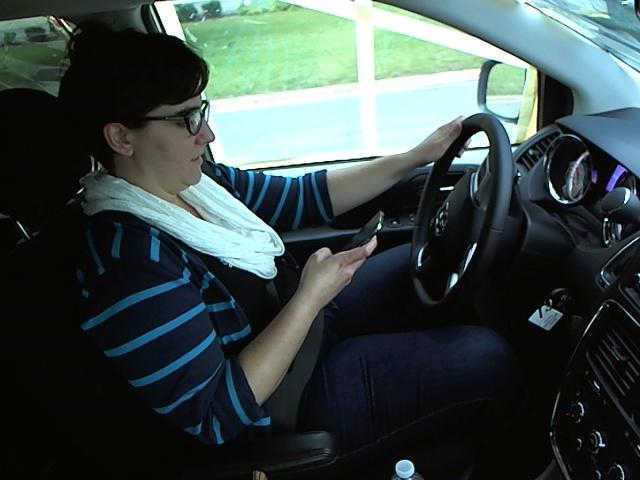}%
\label{Fig. SFDDD 3}}
\hfill
\subfloat[SFDDD texting - right 4]{\includegraphics[width=0.25\linewidth]{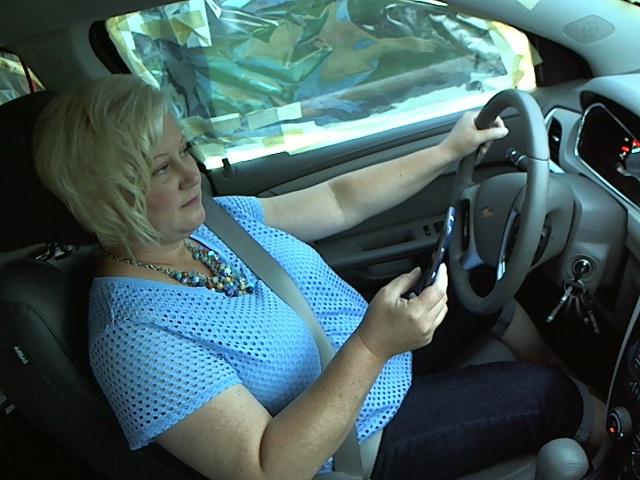}%
\label{Fig. SFDDD 4}}
\\
\subfloat[DriveAct magazine 1]{\includegraphics[width=0.25\linewidth]{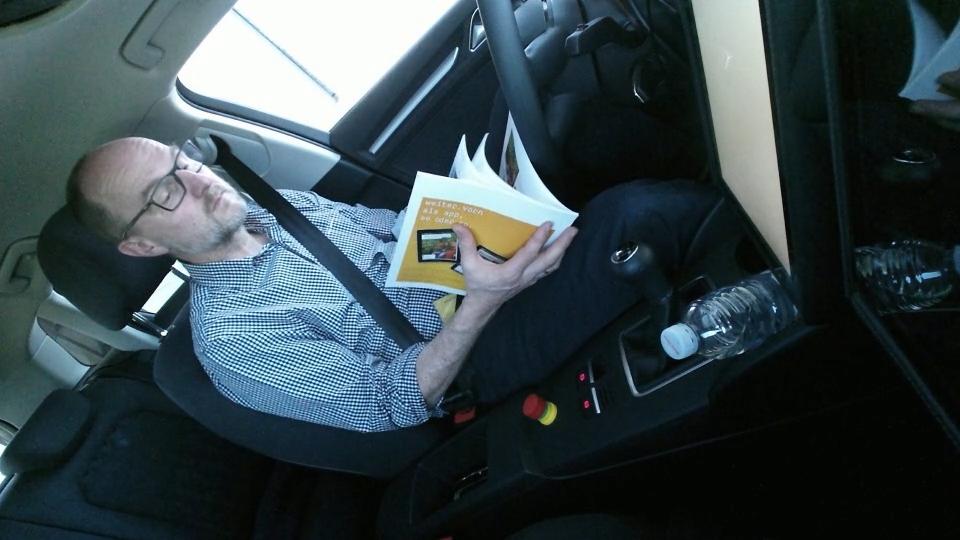}%
\label{Fig. DriveAct 1}}
\hfill
\subfloat[DriveAct magazine 2]{\includegraphics[width=0.25\linewidth]{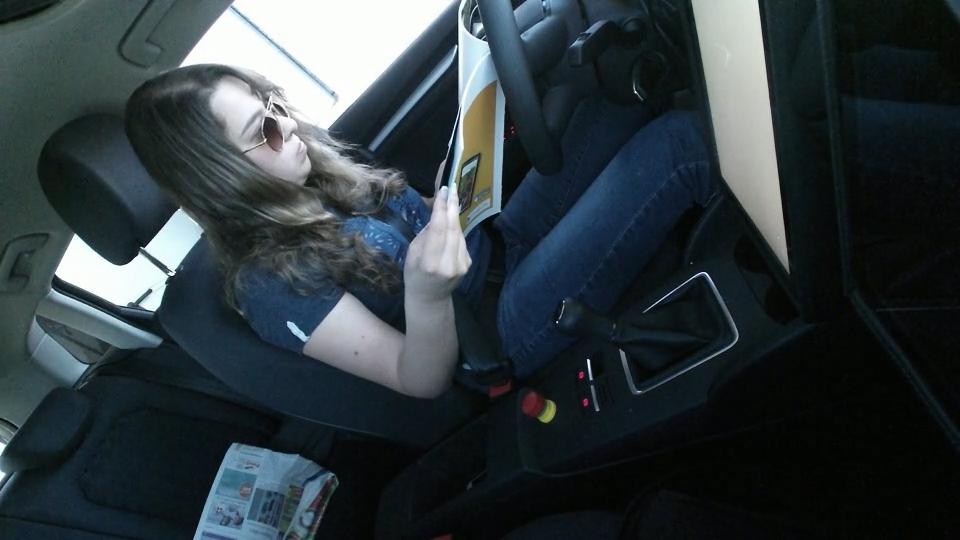}%
\label{Fig. DriveAct 2}}
\subfloat[DriveAct magazine 3]{\includegraphics[width=0.25\linewidth]{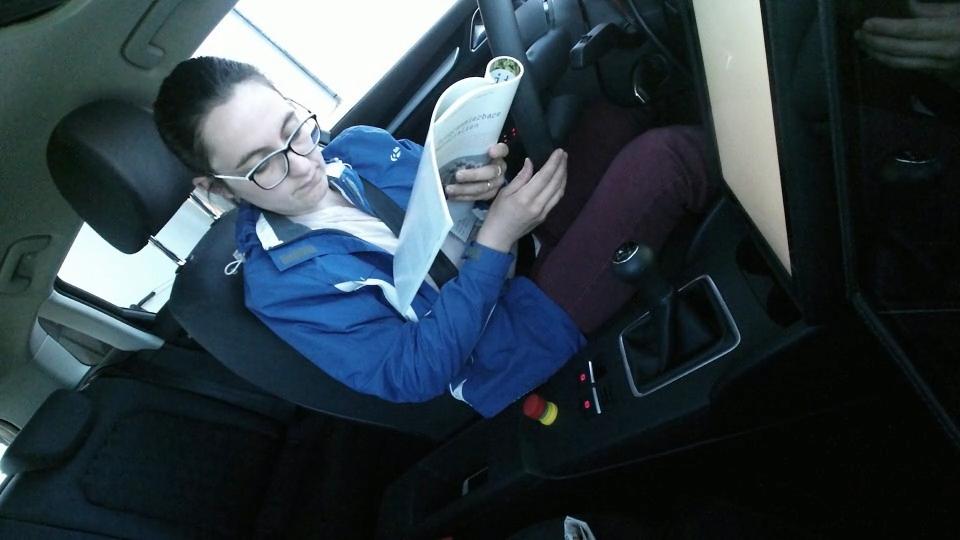}%
\label{Fig. DriveAct 3}}
\hfill
\subfloat[DriveAct magazine 4]{\includegraphics[width=0.25\linewidth]{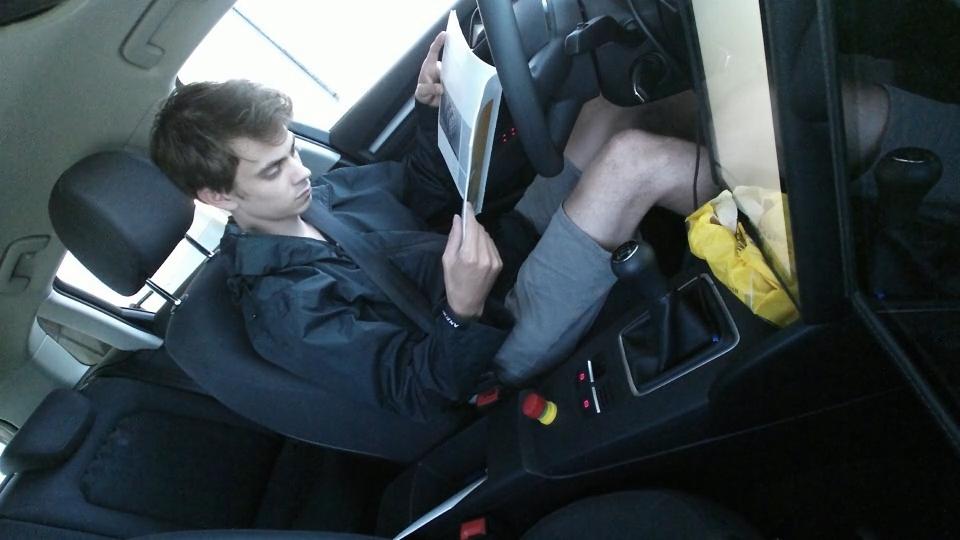}%
\label{Fig. DriveAct 4}}
\caption{Exampled activities of four drivers in each of SFDDD and DriveAct datasets.}
\label{Fig. SFDDD and DriveAct dataset} 
\end{figure*}

\begin{figure}[t]
\centering
\subfloat[SFDDD]{\includegraphics[width=0.5\linewidth]{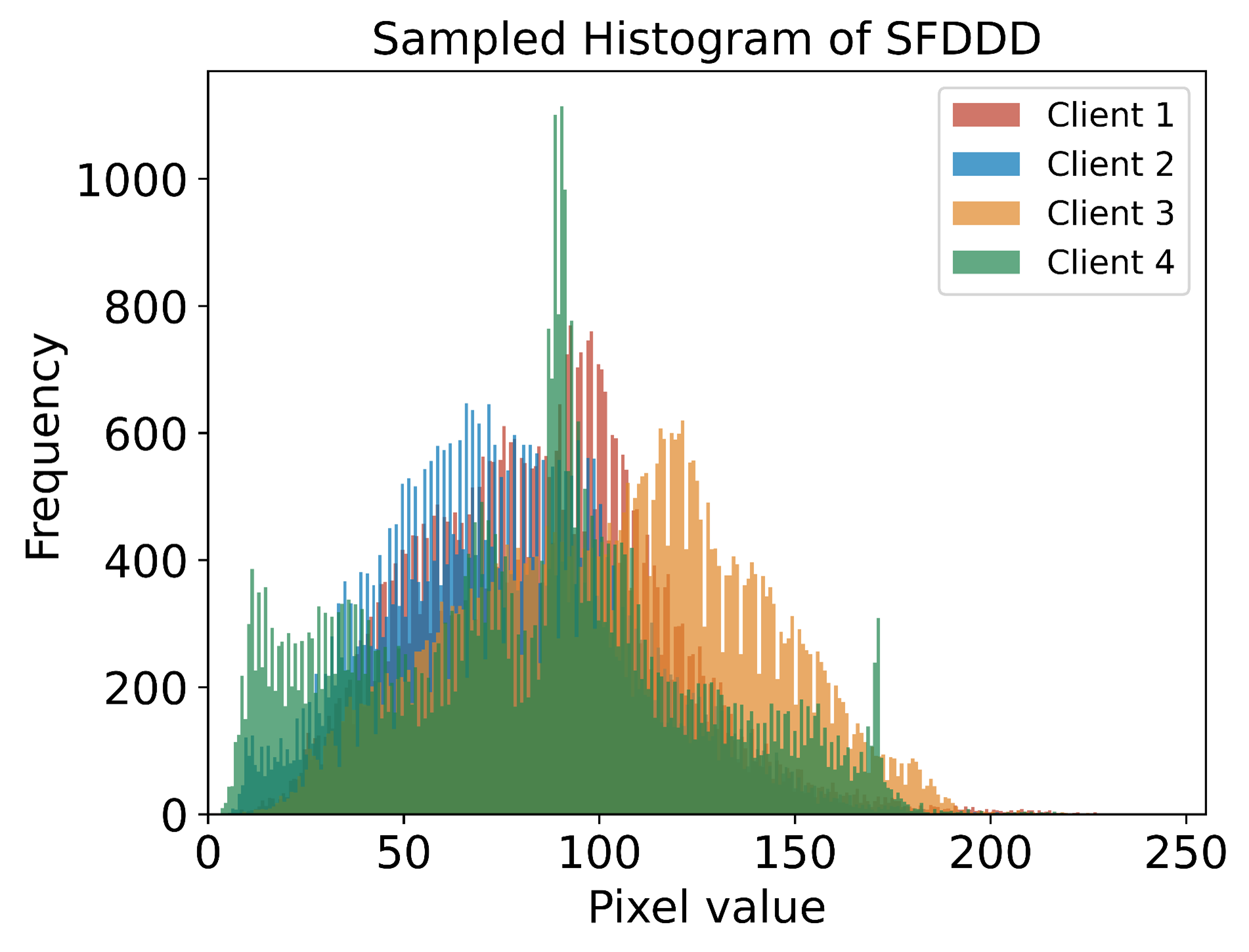}%
\label{Fig. Histogram of SFDDD}}
\hfill
\subfloat[DriveAct]{\includegraphics[width=0.5\linewidth]{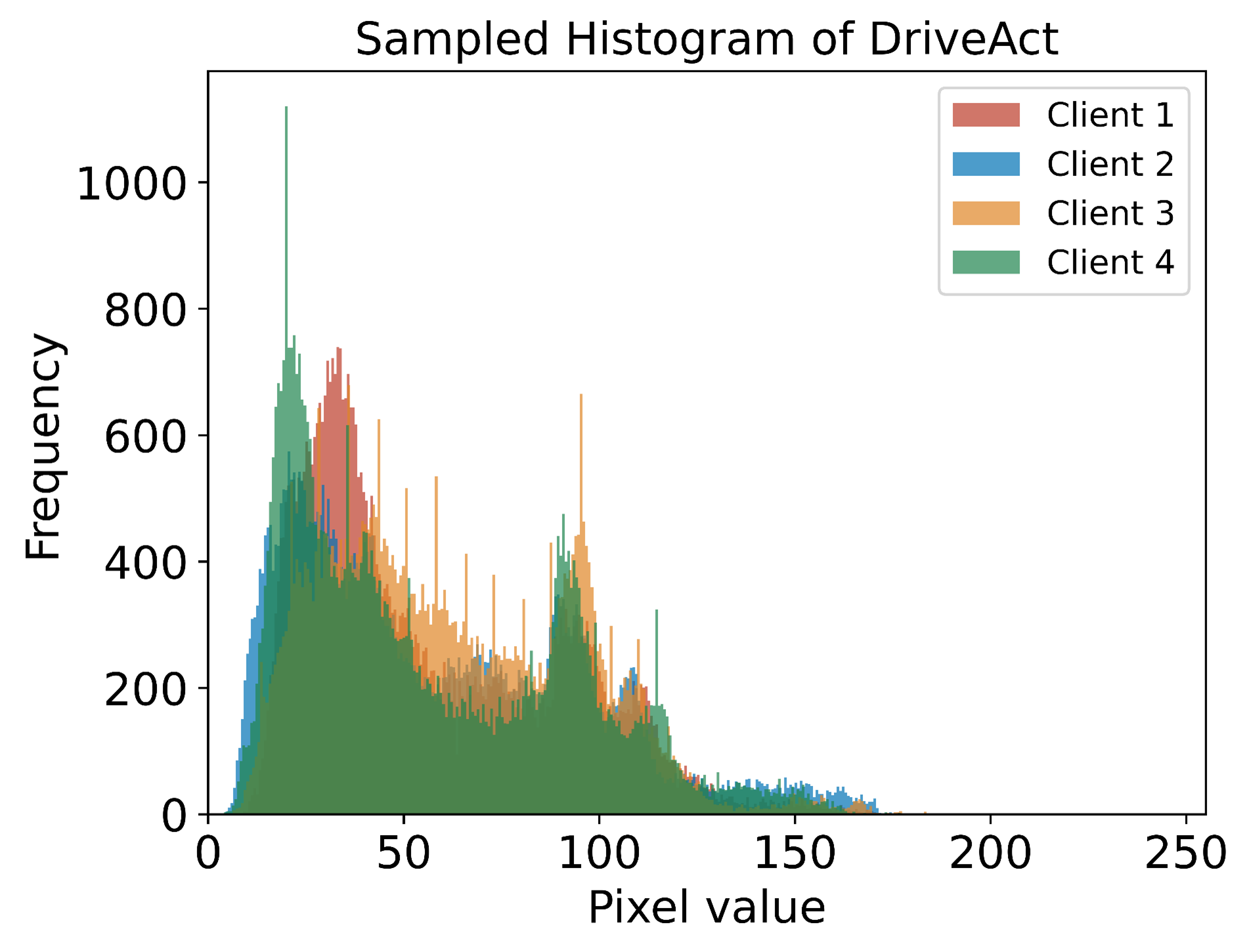}%
\label{Fig. Histogram of DriveAct}}
\caption{Sampled client image histograms of (a) SFDDD and (b) DriveAct datasets.}
\label{Fig. Histogram} 
\end{figure}

\section{Experiment and Results}
\label{Sec. Experiment and Results}

Considering the data and system heterogeneity, experiments are conducted on two open real-world driver monitoring datasets, including State Farm Distracted Driver Detection (SFDDD) \cite{farm_2016} and DriveAct \cite{martin2019drive}. In addition to comparing with FedProx as a baseline, this paper also compares the performance of the transfer, ordered, and personalized extensions through an ablation study.

\subsection{Experiment Setup}
\label{Sec. Experiment Setup}

To compare the impact of system heterogeneity on FL frameworks, the proposed FedTOP is tested on driver monitoring datasets with and without system heterogeneity. The SFDDD dataset includes 26 drivers and 10 activities, and the DriveAct dataset includes 15 drivers and 12 activities. The SFDDD dataset considers system heterogeneity, that is, different drivers have different vehicles, different seat positions, different camera angles, etc., as shown in Fig. \ref{Fig. SFDDD 1}, \ref{Fig. SFDDD 2}, \ref{Fig. SFDDD 3}, and \ref{Fig. SFDDD 4}. The DriveAct dataset does not take into account system heterogeneity, i.e., all subjects had their data collected in the same system. Recorded from the same camera angle, different drivers read the same magazine in the same vehicle, as shown in Fig. \ref{Fig. DriveAct 1}, \ref{Fig. DriveAct 2}, \ref{Fig. DriveAct 3}, and \ref{Fig. DriveAct 4}. 

To show more clearly and visually the heterogeneity between different clients in the two datasets, Fig. \ref{Fig. Histogram} shows histograms of the sample images of the two datasets. It can be seen that the SFDDD dataset with system heterogeneity has a more considerable difference in the distribution of histograms than the DriveAct dataset without system heterogeneity, and the mean value of the SFDDD images is larger. The possible reason is that the vehicle interiors of the DriveAct dataset view are darker, resulting in most of the pixel values being lower. Therefore, FL frameworks may be more challenged by the scene information when training on the SFDDD dataset, such as different vehicle interiors.

Clients are naturally divided based on the drivers. In order to better demonstrate the role of personalized-extension, the datasets are first divided into training clients and testing clients at a ratio of about 0.8, 0.2, with $|C_{\text{SFDDD}}|=20$, $|C'_{\text{SFDDD}}|=6$, $|C_{\text{DriveAct}}|=12$, and $|C'_{\text{DriveAct}}|=3$. And then, the datasets for each client are divided into a training set, verification set, and testing set at a ratio of 0.7, 0.15, and 0.15, respectively. After the global model is trained by the training dataset of training clients, the final trained global model is shared with all clients for personalization. The personalization of the global model will only be processed on the training sets, while the personalized local model will be tested on the unseen testing sets. FL architectures are established on Pytorch and trained on an Intel(R) Core(TM) i9-10850K CPU @ 3.60GHz, and a Nvidia GeForce RTX(TM) 3080 GPU.



\subsection{Ablation Study and Results}
\label{Sec. Ablation Study and Results}
We explore the role of each FedTOP extension on two real-world datasets through an ablation study. FedProx is used as a baseline for comparison. According to the experimental setup described in the previous subsection, the experimental results are shown in Table \ref{Table Experimental results}. 
\begin{table*}[ht] 
\caption{Performance of FedTOP and ablation study on SFDDD and DriveAct datasets.}
\label{Table Experimental results}
\centering
\begin{tabular}{llccccccccc}
\hline
Dataset & Method \textsuperscript{1} & $|C|$ & $q$ & $\mu$ & Transfer &\multicolumn{2}{c}{Accuracy ($\%$) \textsuperscript{2}} & $\text{Time}_\downarrow$ ($\%$) \textsuperscript{3} & $\text{Commun}_\downarrow$ ($\%$) \textsuperscript{4} & Cybersecurity \\
~       & ~     & ~   & ~     & ~     & ~     & Training & Testing & ~ & ~ & ~ \\
\hline
SFDDD & FedProx (baseline) & 20 & 20 & 1 & No & 54.63 & 16.44 & $\sim$ & $\sim$ & $\sim$ \\
~     & FedOP & 20 & 15 & 1 & No & 97.69 & 96.37 & 1.45 $\downarrow$ & $\sim$ & $\uparrow$ \\
~     & FedTP & 20 & 20 & 1 & Yes & 94.76 & 92.8 & 17.3 $\downarrow$ & 37.46 $\downarrow$ & $\sim$ \\
~     & FedTO & 20 & 15 & 1 & Yes & 46.16 & 16.43 & 18.91 $\downarrow$ & 37.46 $\downarrow$ & $\uparrow$ \\
~     & \textbf{FedTOP} & \textbf{20} & \textbf{15} & \textbf{1} & \textbf{Yes} & \textbf{94.65} & \textbf{92.32} & \textbf{18.91} $\boldsymbol{\downarrow}$ & \textbf{37.46} $\boldsymbol{\downarrow}$ & \text{ } $\boldsymbol{\uparrow}$ \\
DriveAct   & FedProx (baseline) & 12 & 12 & 1 & No  & 73.18 & 23.96 & $\sim$ & $\sim$ & $\sim$ \\
~          & FedOP & 12 & 10 & 1 & No & 98.07 & 97.97 & 0.44 $\downarrow$ & $\sim$ & $\uparrow$ \\
~          & FedTP & 12 & 12 & 1 & Yes & 97.00 & 95.71 & 16.83 $\downarrow$ & 37.46 $\downarrow$ & $\sim$ \\
~          & FedTO & 12 & 10 & 1 & Yes & 62.30 & 22.89 & 19.18 $\downarrow$ & 37.46 $\downarrow$ & $\uparrow$ \\
~          & \textbf{FedTOP} & \textbf{12} & \textbf{10} & \textbf{1} & \textbf{Yes} & \textbf{97.04} & \textbf{95.96} & \textbf{19.18} $\boldsymbol{\downarrow}$ & \textbf{37.46} $\boldsymbol{\downarrow}$ & \text{ } $\boldsymbol{\uparrow}$ \\
\hline
\end{tabular}

\noindent{\textsuperscript{1} FedOP, FedTP, and FedTO refer to ablating the transfer, ordered, and personalized extensions of the FL framework, respectively. FedOP refers to training and transmitting the entire model without using the transfer learning paradigm. FedTP refers to the aggregation of all models without ordered dropout of larger loss models. FedTO refers to the use of the global model without local personalization.}
\noindent{\textsuperscript{2} Accuracy refers to the testing sets of training clients and testing clients, which is described in Section \ref{Sec. Experiment Setup}.}
\noindent{\textsuperscript{3} $\text{Time}_\downarrow$ refers to the ratio of reduced training time per client to the baseline.}
\noindent{\textsuperscript{4} $\text{Commun}_\downarrow$ refers to the ratio of reduced communication consumption to the baseline, which is described in (\ref{Eq. communication resources}).}
\end{table*}

\begin{figure*}[t]
\centering
\subfloat[FedProx]{\includegraphics[width=0.25\linewidth]{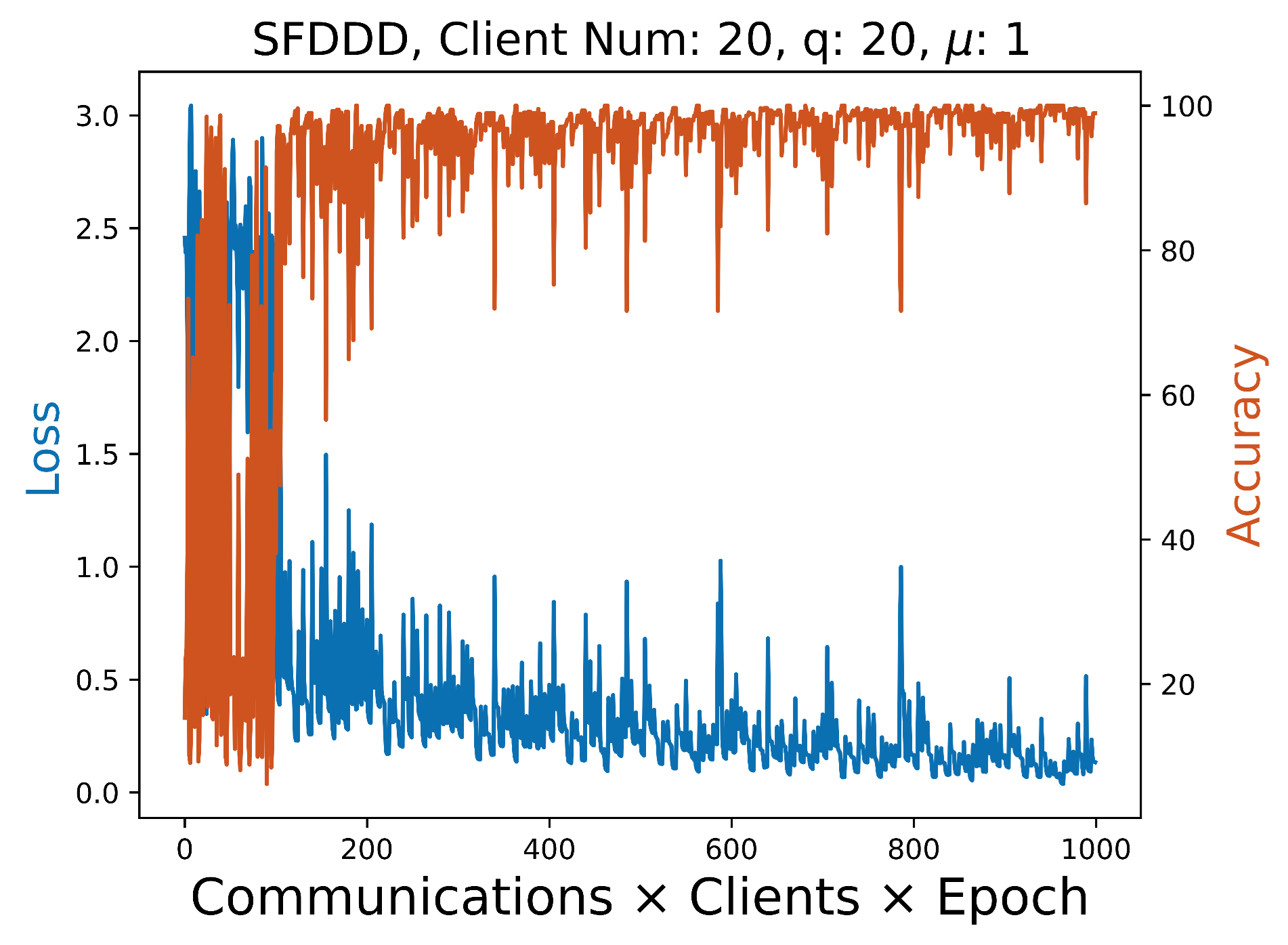}%
\label{Fig. SFDDD FedProx}}
\hfill
\subfloat[FedO]{\includegraphics[width=0.25\linewidth]{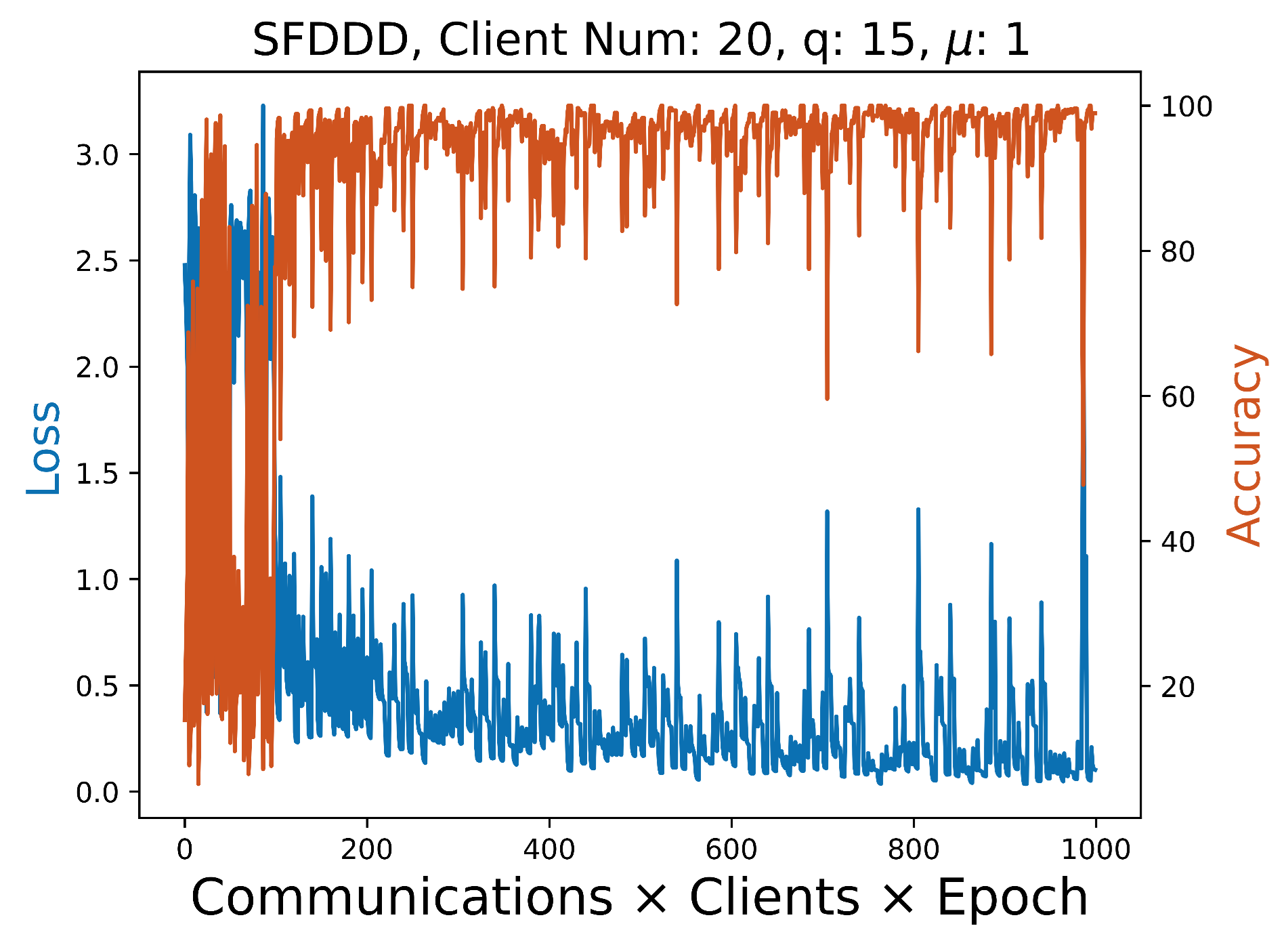}%
\label{Fig. SFDDD OP}}
\hfill
\subfloat[FedT]{\includegraphics[width=0.25\linewidth]{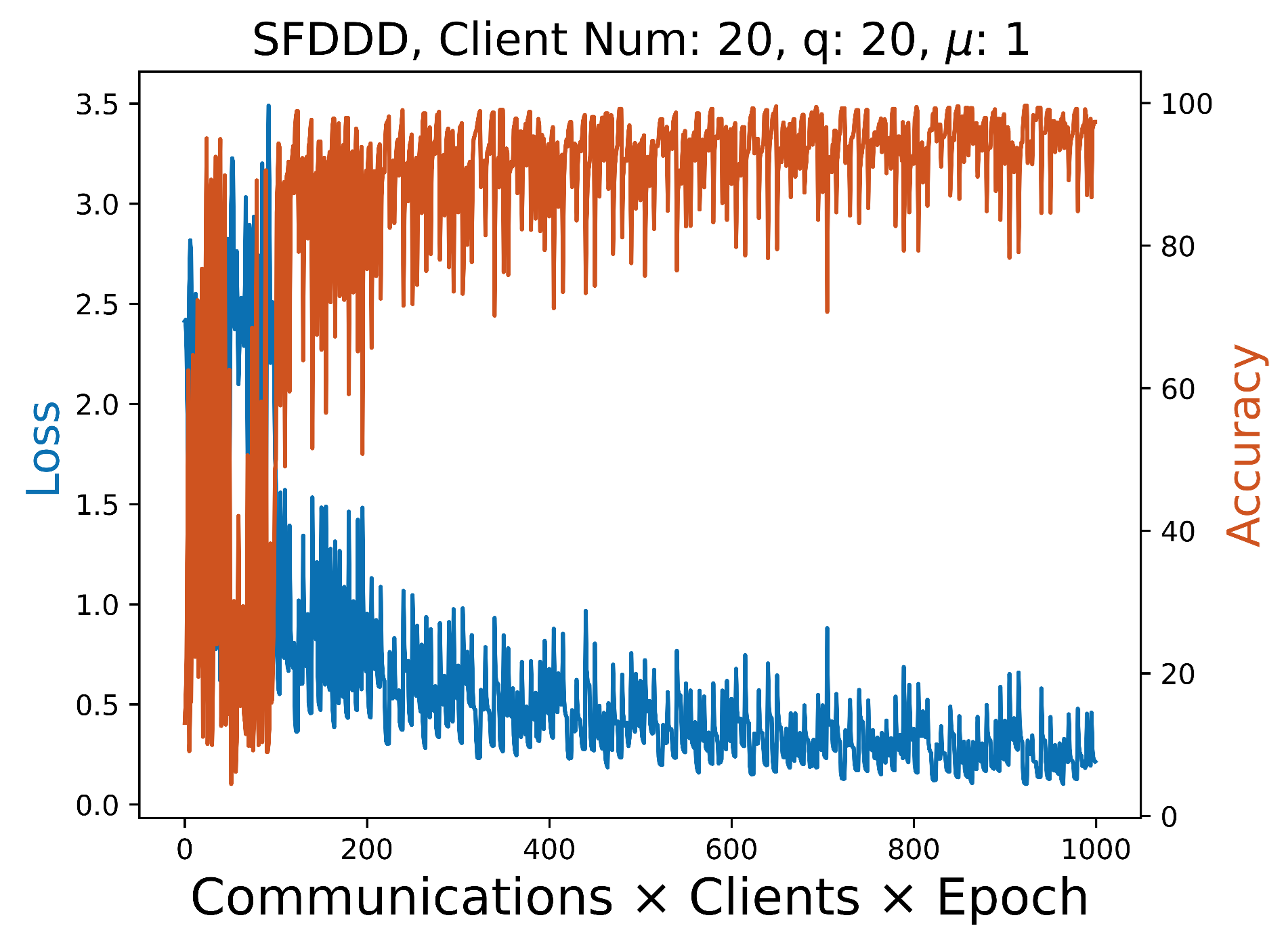}%
\label{Fig. SFDDD TP}}
\hfill
\subfloat[FedTO]{\includegraphics[width=0.25\linewidth]{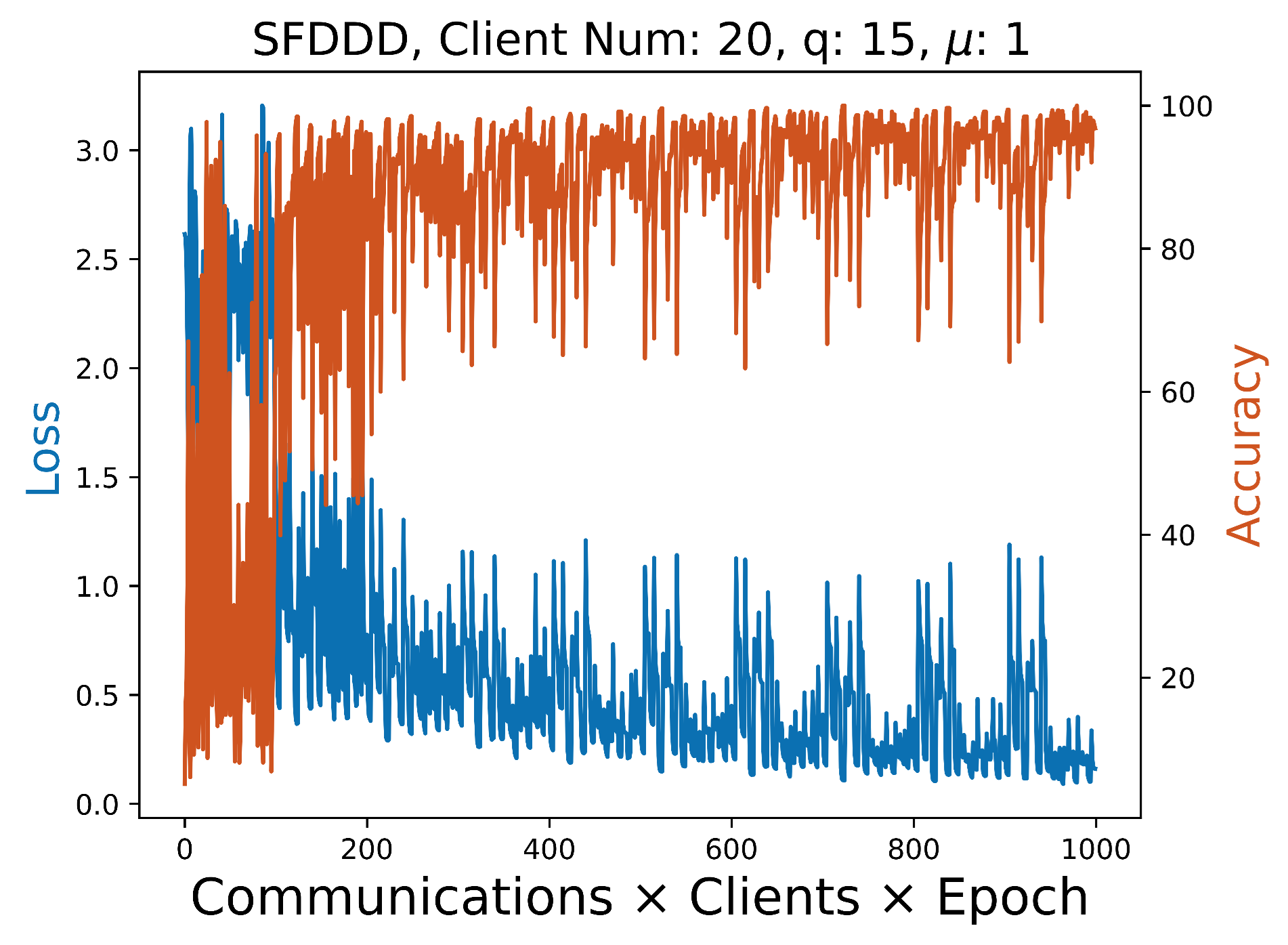}%
\label{Fig. SFDDD TO}}
\\
\subfloat[FedProx]{\includegraphics[width=0.25\linewidth]{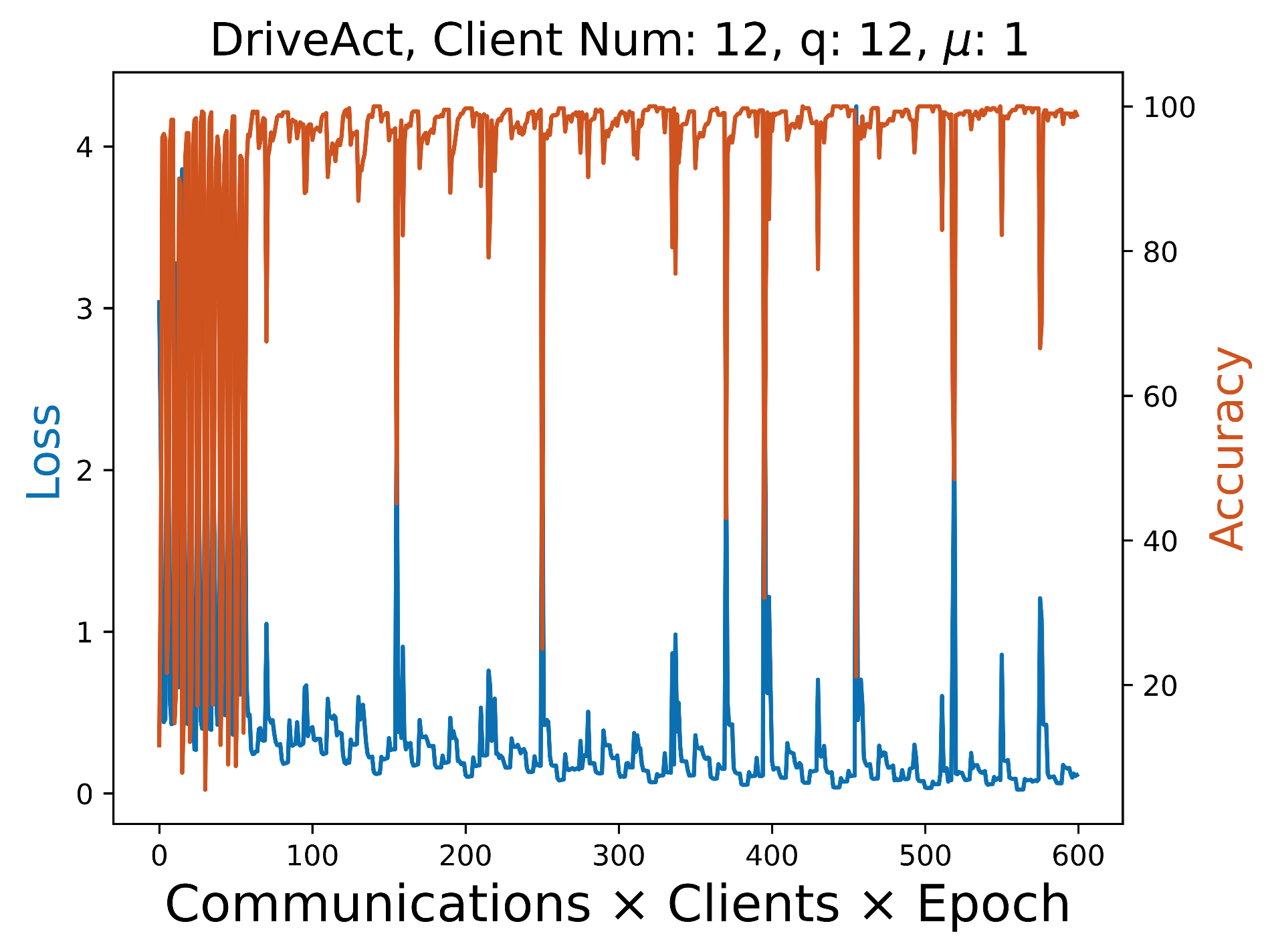}%
\label{Fig. DriveAct FedProx}}
\hfill
\subfloat[FedO]{\includegraphics[width=0.25\linewidth]{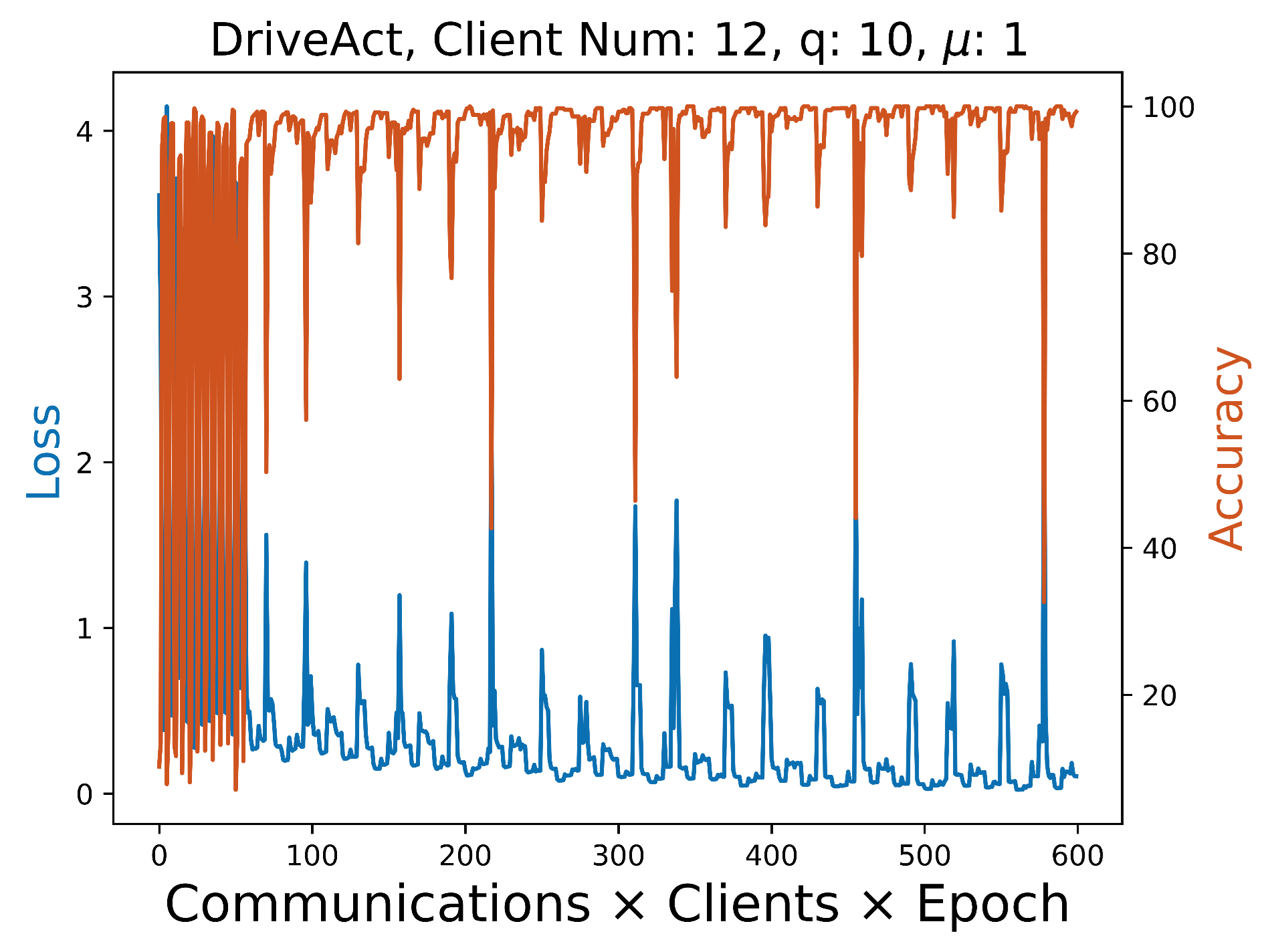}%
\label{Fig. DriveAct OP}}
\hfill
\subfloat[FedT]{\includegraphics[width=0.25\linewidth]{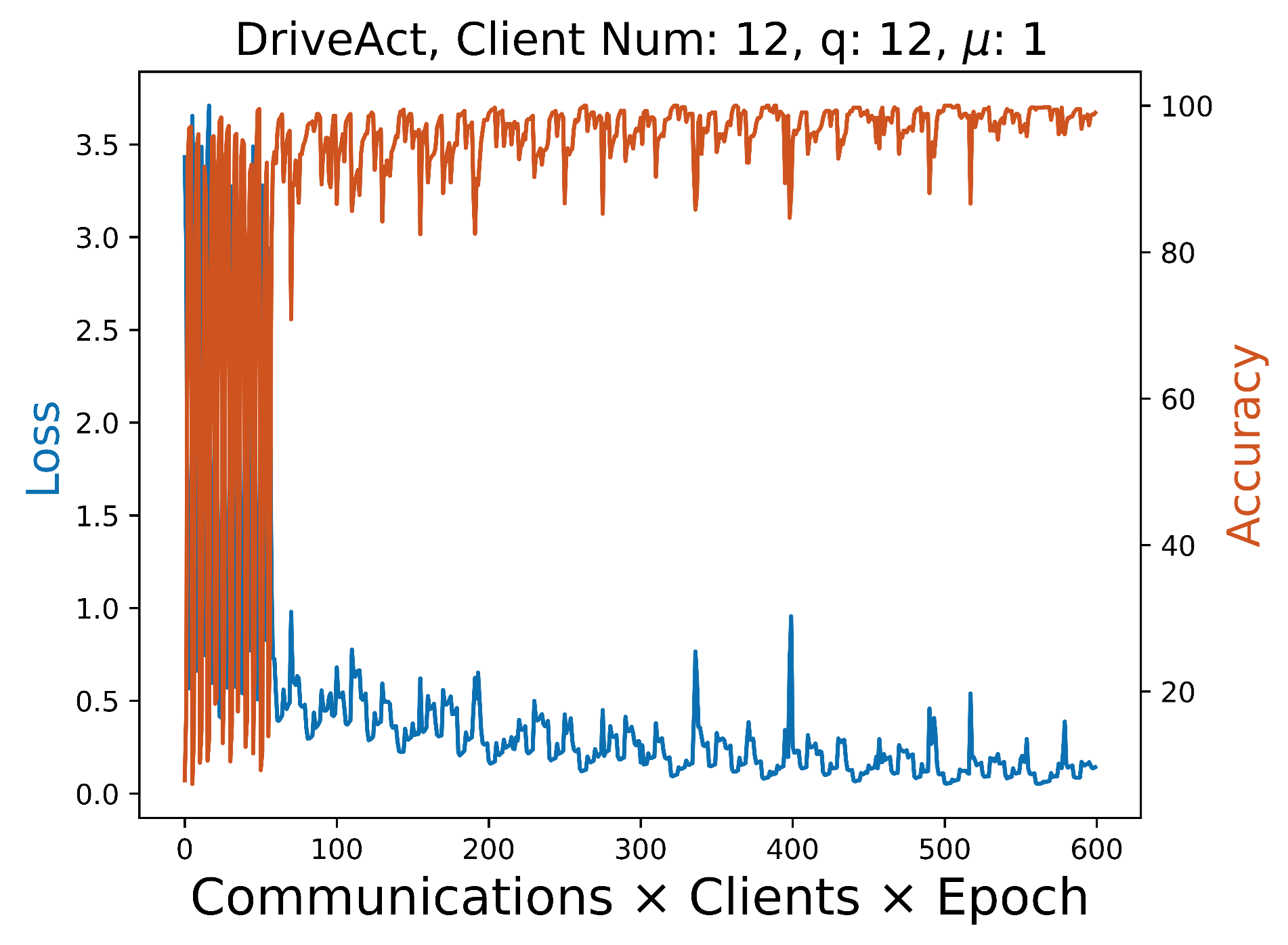}%
\label{Fig. DriveAct TP}}
\hfill
\subfloat[FedTO]{\includegraphics[width=0.25\linewidth]{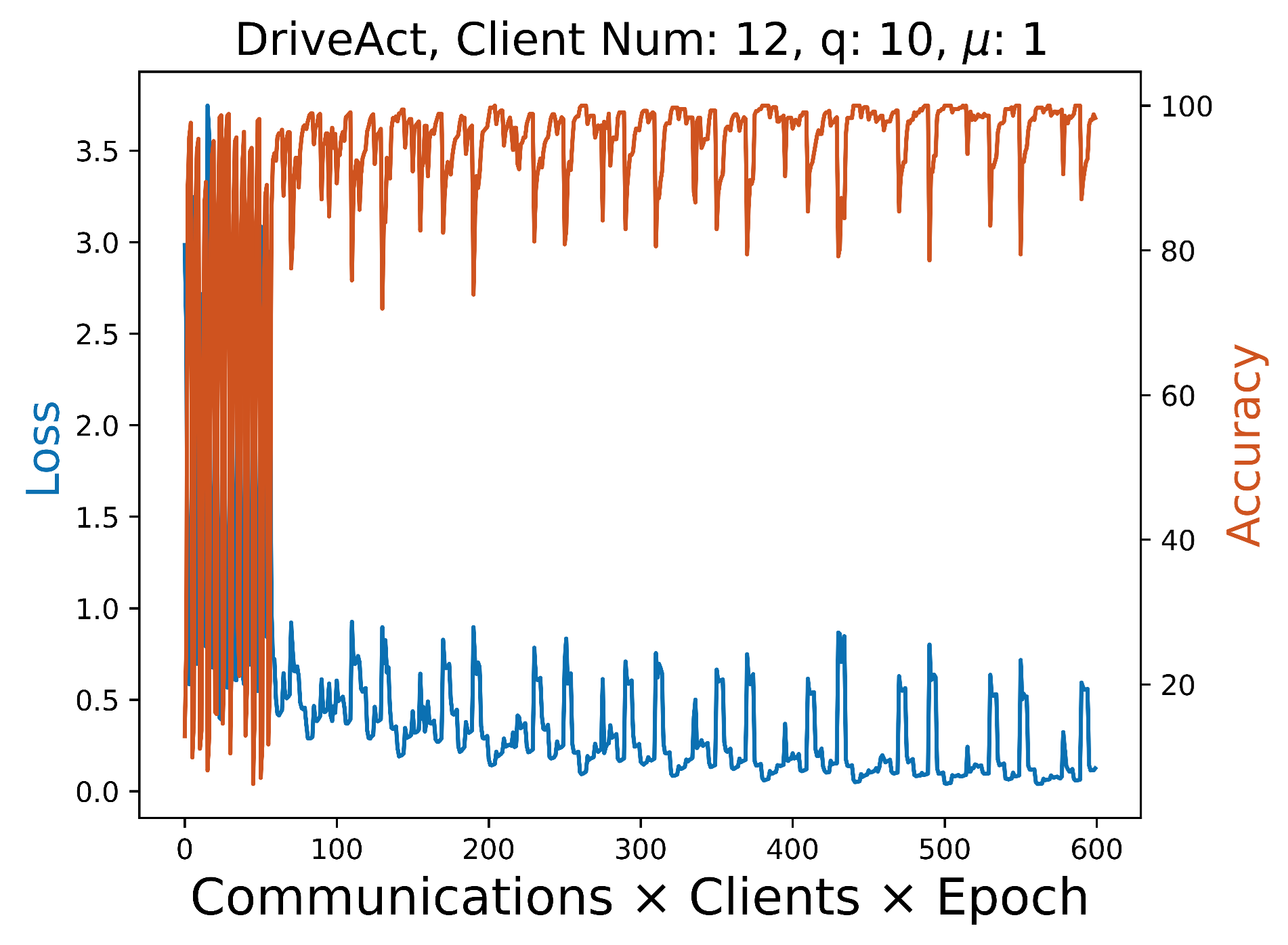}%
\label{Fig. DriveAct TO}}
\caption{Accuracy and loss curves of the FL framework and its extensions on the SFDDD and DriveAct datasets, which is the training process of Algorithm \ref{Alg FedTOP}. Note that personalization does not affect the convergence of the global model in FL frameworks, i.e., FedTO converges the same as FedTOP. FedO and FedT refer to ablating the transfer and ordered extensions of the FedTOP framework, respectively.}
\label{Fig. SFDDD and DriveAct Results} 
\end{figure*}

\begin{figure}[t]
\centering
\subfloat[SFDDD, FedTOP]{\includegraphics[width=0.5\linewidth]{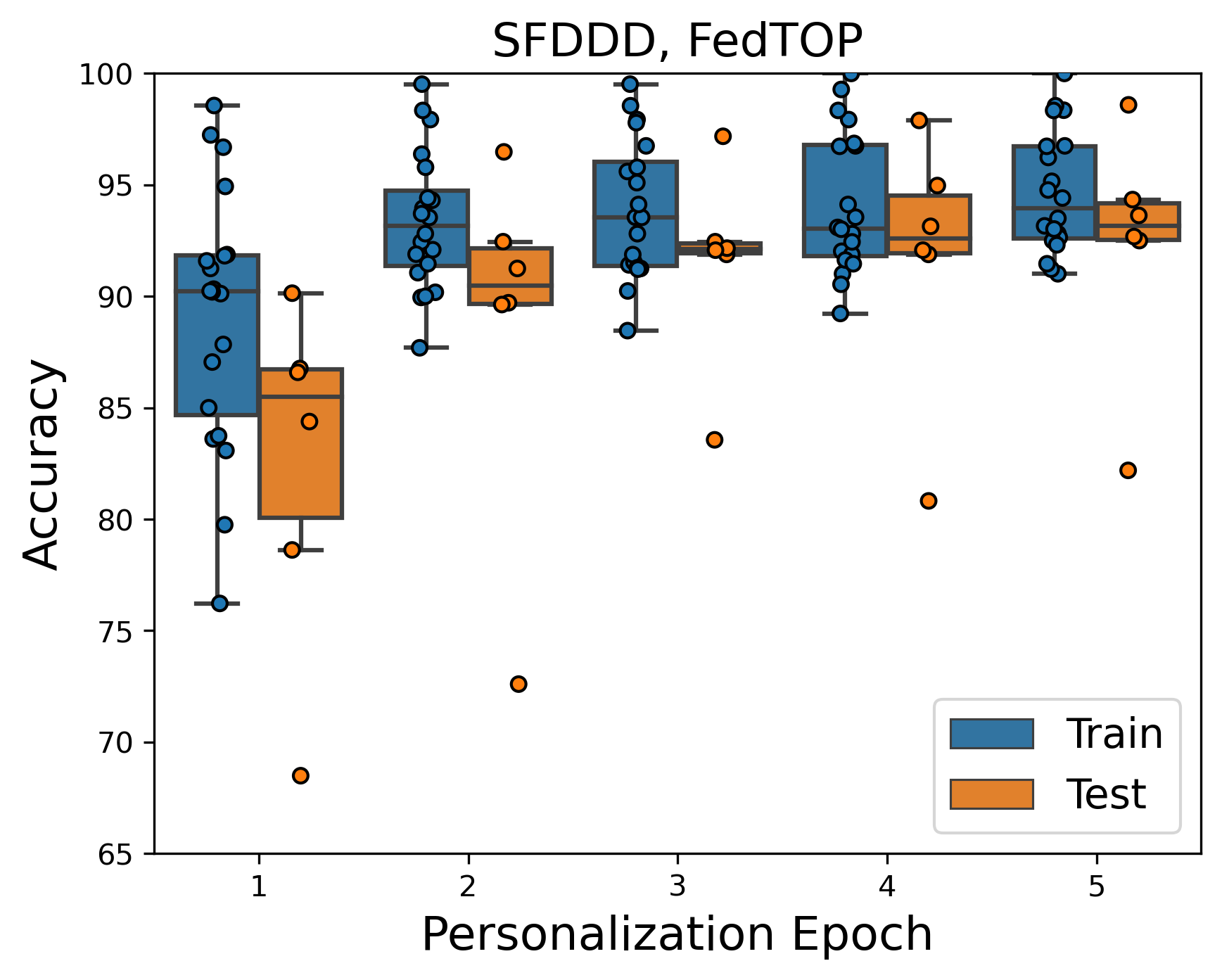}%
\label{Fig. SFDDD TOP}}
\hfill
\subfloat[DriveAct, FedTOP]{\includegraphics[width=0.5\linewidth]{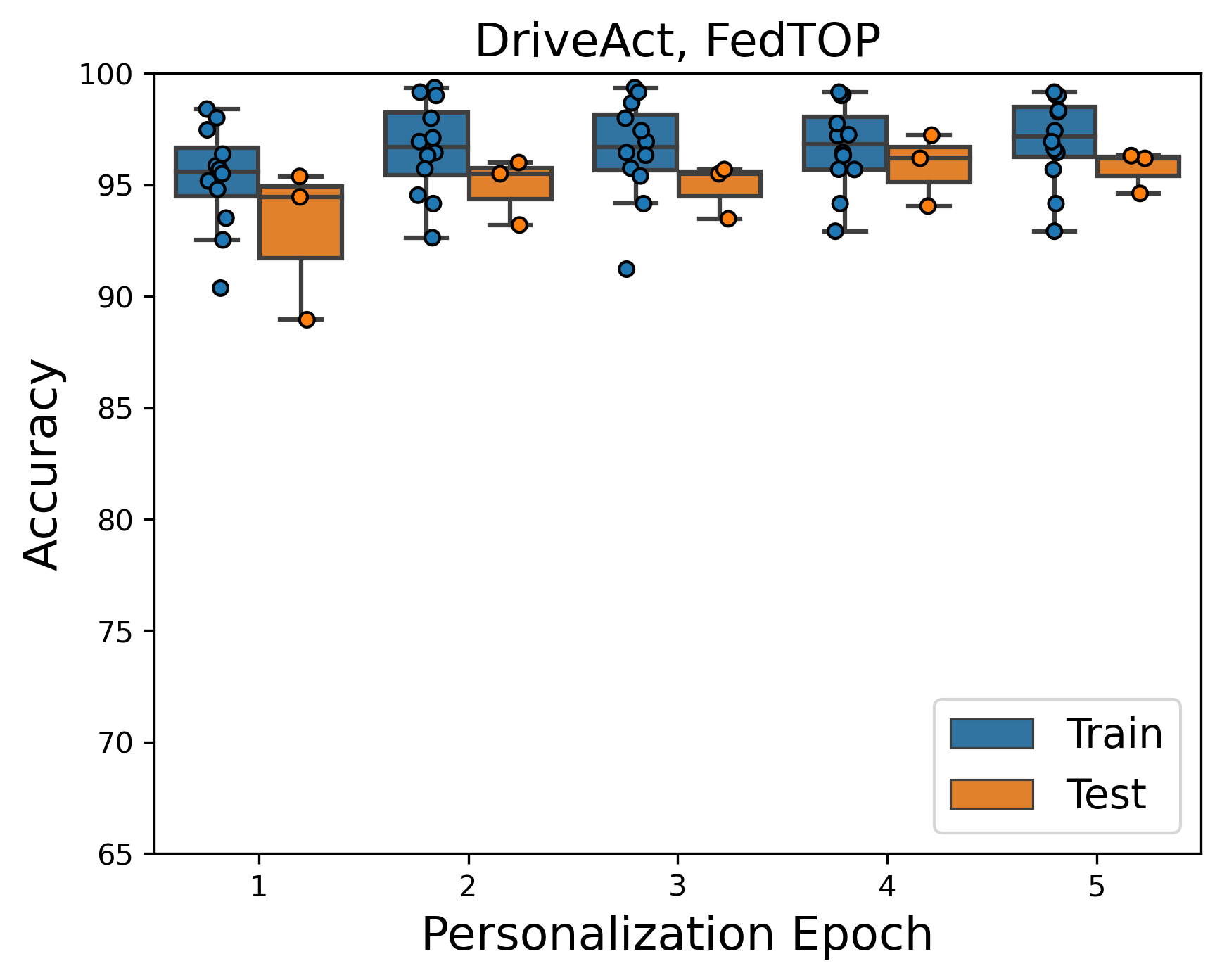}%
\label{Fig. DriveAct TOP}}
\caption{Error bars and accuracy distributions for training and testing clients on (a) SFDDD and (b) DriveAct datasets. Each scatter represents a client. Testing accuracy of the training and testing clients varies with personalized epoch, which is the results of Algorithm \ref{Alg Personalization}.}
\label{Fig. Personalization Results} 
\end{figure}

The results and comparisons for two datasets and three extensions are shown in Fig. \ref{Fig. SFDDD and DriveAct Results}, which is equivalent to demonstrating Algorithm \ref{Alg FedTOP}. By observing the accuracy and loss curves on the two datasets, it can be concluded that the SFDDD dataset with system heterogeneity is fundamentally different from the DriveAct dataset without system heterogeneity. It can be clearly seen that the SFDDD dataset with system heterogeneity requires more communication to converge, while the DriveAct dataset without system heterogeneity has a fast convergence speed, especially at the first communication. Therefore, for real-world datasets, system heterogeneity can be mitigated by more communication times. 

\begin{figure*}[t]
\centering
\subfloat[Trained global model $\omega^T$]{\includegraphics[width=0.25\linewidth]{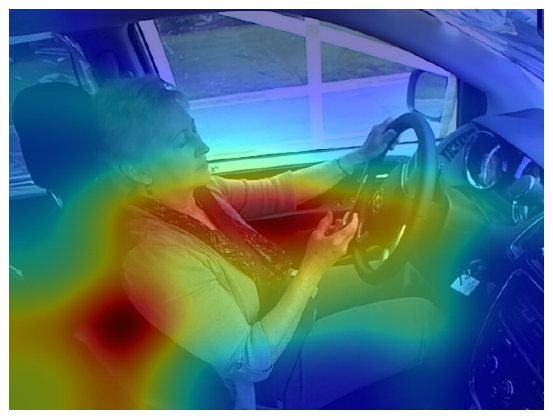}%
\label{Fig. SFDDD P 0}}
\hfill
\subfloat[Personalization Epoch 1 $\omega^{T^1}$]{\includegraphics[width=0.25\linewidth]{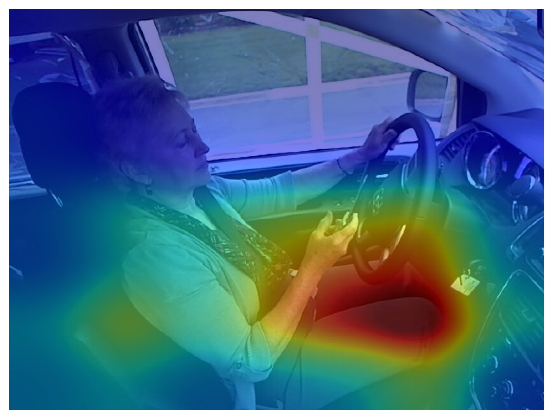}%
\label{Fig. SFDDD P 1}}
\hfill
\subfloat[Personalization Epoch 3 $\omega^{T^3}$]{\includegraphics[width=0.25\linewidth]{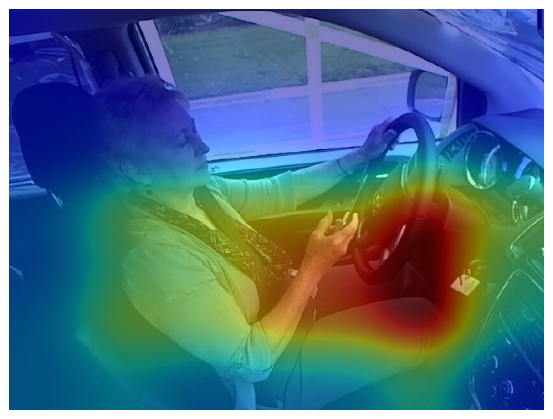}%
\label{Fig. SFDDD P 3}}
\hfill
\subfloat[Personalization Epoch 5 $\omega^{T^5}$]{\includegraphics[width=0.25\linewidth]{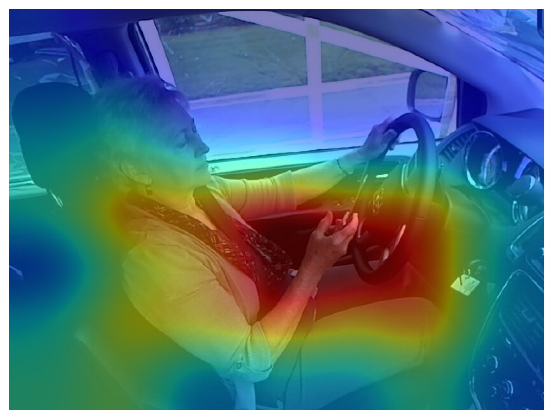}%
\label{Fig. SFDDD P 5}}
\\
\subfloat[Trained global model $\omega^{T}$]{\includegraphics[width=0.25\linewidth]{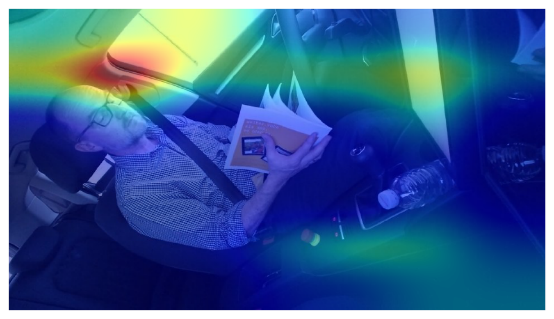}%
\label{Fig. DriveAct P 0}}
\hfill
\subfloat[Personalization Epoch 1 $\omega^{T^1}$]{\includegraphics[width=0.25\linewidth]{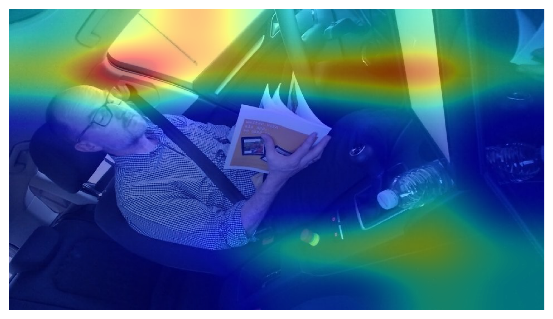}%
\label{Fig. DriveAct P 1}}
\hfill
\subfloat[Personalization Epoch 3 $\omega^{T^3}$]{\includegraphics[width=0.25\linewidth]{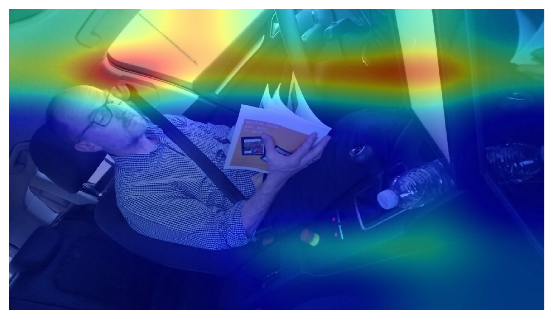}%
\label{Fig. DriveAct P 3}}
\hfill
\subfloat[Personalization Epoch 5 $\omega^{T^5}$]{\includegraphics[width=0.25\linewidth]{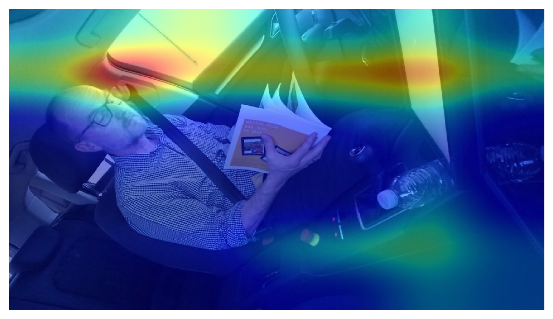}%
\label{Fig. DriveAct P 5}}
\\
\caption{CAMs of the test clients in SFDDD and DriveAct datasets during the personalization process. (a), (b), (c), and (d) are test clients in the SFDDD dataset, which is the same as Fig. \ref{Fig. SFDDD 1}. (e), (f), (g), and (h) are test clients in the DriveAct dataset, which is the same as Fig. \ref{Fig. DriveAct 1}.}
\label{Fig. Class activation map} 
\end{figure*}

By observing Fig. \ref{Fig. SFDDD OP}, \ref{Fig. SFDDD TO}, \ref{Fig. DriveAct OP}, and \ref{Fig. DriveAct TO},  it can be found that the ordered-extension diminishes the stability of the system. Although the anomalous large-loss local model is discarded to reduce the bias of the global model, it also increases the variance of the global model resulting in reduced generalizability. By observing Fig. \ref{Fig. SFDDD TP}, \ref{Fig. SFDDD TO}, \ref{Fig. DriveAct TP}, and \ref{Fig. DriveAct TO}, we can see that the effect of transfer-extension is different for datasets with and without system heterogeneity. On the one hand, transfer-extension increases the variance of the model on the SFDDD dataset and leads to a reduced and unstable model convergence. On the other hand, transfer-extension improves the speed of model convergence on DriveAct, and the convergence effect is more stable. The possible reason is that the transfer-extension retains only a small number of trainable parameters, resulting in the neural network model not being able to learn human behavioral features effectively in the SFDDD dataset with system heterogeneity. However, for the DriveAct dataset without system heterogeneity, the factors are constant except for the driver, and the local model does not need to focus on these exact same pixels, but only on the changing pixels, including objects such as drivers, computers, and magazines. Therefore, for the DriveAct dataset, transfer-extension can effectively increase convergence and stability. The proposed FedTOP framework is able to obtain 92.32$\%$ and 95.96$\%$ accuracy on the SFDDD and DriveAct datasets, respectively, when considering five times of personalization training. Compared to FedProx as a baseline, FedTOP can effectively improve the accuracy by 462$\%$ in addition to considering a 37.46$\%$ reduction in communication resources. The results demonstrate the feasibility of the proposed FedTOP in terms of communication resource-saving, accuracy improvement, robustness, and cybersecurity.

\subsection{Performance of Personalized-Extension}
\label{Sec. Performance of Personalized-extension}
Personalized-extension needs to be further discussed and analyzed as the most effective approach to improve accuracy. Based on the division of training and testing clients in Section \ref{Sec. Experiment Setup}, in this subsection, we further discuss how the trained and aggregated global model is adapted to both training and testing clients. The results of the personalized-extension on the two datasets are shown in Fig. \ref{Fig. Personalization Results} with different personalization epochs, which is equivalent to demonstrating Algorithm \ref{Alg Personalization}. It can be seen that the personalization process differs significantly on the datasets with and without system heterogeneity, which is similar to the results in Fig. \ref{Fig. SFDDD and DriveAct Results}. The clients in the DriveAct dataset have faster convergence, minor accuracy variance, and higher final accuracy. On the contrary, the clients in the SFDDD dataset not only converge slower but also have an anomalous client with relatively low accuracy. The possible reason is that the anomalous client has a huge data and system heterogeneity, causing the optimal model to deviate significantly far from the aggregated global model.

Fig. \ref{Fig. Class activation map} further demonstrates that the trained global model repositions the ROI during the personalized training process via class activation map (CAM) \cite{omeiza2019smooth}. The test client of the SFDDD dataset can be seen struggling with the personalization process. The trained global model focuses the ROI on the seat backrest, driver's chest, hand, and knee, and vehicle door. Due to the system heterogeneity present in the SFDDD dataset, the positions of the driver, seat, and steering wheel, as shown in Fig. \ref{Fig. SFDDD P 0} is different from other clients, as shown in Fig. \ref{Fig. SFDDD 2}, \ref{Fig. SFDDD 3}, and \ref{Fig. SFDDD 4}. Therefore, the initial ROI is likely to be a driver's position among other clients. During the five personalization training processes, the local model is able to effectively reposition the ROI to the driver, which is what the personalized-extension is intended to show. Moreover, the personalization process also reduces the number of ROI while targeting more attention to a specific area. 

On the contrary, for the test clients in the DriveAct dataset, the adjustment of the ROI is negligible. Note that the ROI does not necessarily have to cover the driver's body or an object such as the magazine. The ROI should cover those pixels that can distinguish between different activities, such as static activities like reading the magazine, and dynamic activities like wearing a seatbelt in the DriveAct dataset activity setting. The ROI focuses on areas where large differences are likely to occur. The fact that the ROI in the DriveAct dataset covers almost the same pixels during the personalization process can also prove the negative impact of system heterogeneity on the FL framework.

\section{Discussion}
\label{Sec. Discussion}
The proposed FedTOP is tested on two datasets: SFDDD and DriveAct. Although both datasets involve driver activity classification tasks, they have different objectives. The SFDDD dataset aims to detect driver distraction under normal driving conditions, while the DriveAct dataset is focused on activity recognition in autonomous driving scenarios. The DMA highlighted in this paper is interested in monitoring driver distractions or other risky behaviors, as different activities carry different risk factors. For instance, reaching behind is more hazardous than talking to a passenger. Therefore, in situations where there is a high risk of distracted state or dangerous behavior, the autopilot system can be employed to take control of the vehicle and prevent traffic accidents.

The two datasets used, SFDDD and DriveAct, still have some flaws. First, although the SFDDD dataset takes system heterogeneity into account, quite a few drivers collect data in the same vehicle. Therefore, there are still some differences between the dataset and the real-world data, which leads to the fact that the proposed FedTOP may need more communication rounds to achieve similar accuracy on a real-world dataset. Second, there is currently no driver monitoring dataset with real poisoning data currently existing, resulting in the effect of ordered-extension not being reflected. The different modalities, positions, and angles of the camera or the method of generating fake data may be a hypothesis for poisoned data, but it cannot be highlighted as real. Moreover, due to road safety guidelines, the current dataset is only driving on safe roads or simulated driving. Therefore, the driver's posture, demeanor, facial concentration, etc., are far from the real driving behavior. Therefore, there is an urgent need for a more realistic dataset that can include camera images of different positions and angles, different vehicle scenes, and more drivers driving on real roads.

For a FL framework in IoT, in addition to accuracy being the evaluation criterion, factors like communication requirements, robustness, fairness, cybersecurity, etc., also need to be considered. Although it seems that transfer and ordered extensions may not improve accuracy but rather reduce it in the current experimental results, it can potentially improve the performance of the FL framework. Therefore, we keep two extensions as one of our future directions. Personalized-extension is an approach similar to transfer learning and incremental learning. On the one hand, the local client is incrementally learned based on the trained global model, but it does not intentionally retain the previously learned knowledge. On the other hand, the global model is transferred to the client dataset as in transfer-extension, but the low-level non-trainable weights are still pre-trained on ImageNet. Therefore, the proposed personalized-extension actually uses the trained global model weights to fit different client data, such as the reposition of ROI. Although the personalized-extension requires additional training locally for each client, there are many benefits, including high accuracy, applicability to non-training clients, customization, etc. Conceivably, personalized-extension can effectively address the problem of system heterogeneity, e.g., it can be applied to different cameras, camera angles, vehicle interiors, etc.

\section{Conclusion}
\label{Sec. Conclusion}
In this paper, we propose a FL framework FedTOP for DMA to address the issues of privacy preservation, efficient training, communication resource-saving, poisoned data, and diversified scenarios. Through the ablation study, the impact, role, and performance of three extensions, including transfer, ordered, and personalized on the model, are disclosed. Moreover, the experiments demonstrate dramatic differences between datasets with and without system heterogeneity. In addition to the proposed FedTOP being able to exhibit 92.32$\%$ and 95.96$\%$ accuracy in two datasets for testing clients, it is also appreciated that FedTOP reduces communication consumption by 37.46$\%$ and potentially improves cybersecurity. The experimental results show that the proposed FedTOP is a highly accurate, lightweight, privacy-preserving, robust, cybersecure, and universally applicable FL framework for potential DMAs.

Future work lies in the continued research of extensions. For the ordered-extension, a possible plan is to introduce some malicious local clients to attack and poison with the global model. For example, subjects may not place the camera on the side as instructed but place it on the front or behind instead. Such outliers may cause the global model to deviate significantly from the optimal solution, so in this case, ordered-expansion can prevent the deviation of the global model by discarding the larger value of the losses. For the transfer-extension, there is currently a lack of a general driver monitoring model, so we used a model pre-trained on ImageNet. Future work can pre-train a driver model ourselves as a base model, which will get better performance in DMA. Fig. \ref{Fig. FedTOP structure} shows the FL framework for foresight in IoV, but the dataset used does not contain scenario information such as road, weather, vehicle models, etc. Therefore, we expect a well-developed real-world dataset to include such scenario information, data and system heterogeneity, etc.

\bibliographystyle{IEEEtran}
\small\bibliography{reference}

\vfill

\begin{IEEEbiography}
[{\includegraphics[width=1in,height=1.25in,clip,keepaspectratio]{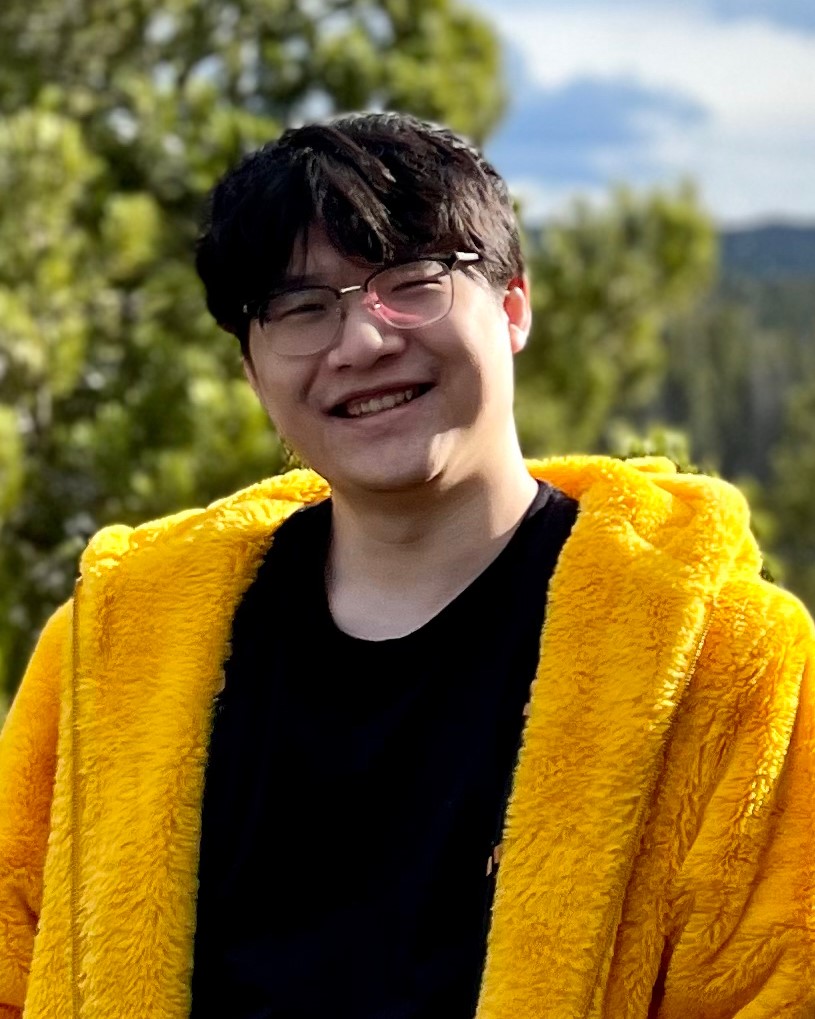}}]
{Liangqi Yuan}
(S'22) received the B.E. degree from the Beijing Information Science and Technology University, Beijing, China, in 2020, and the M.S. degree from the Oakland University, Rochester, MI, USA, in 2022. He is currently pursuing the Ph.D. degree with the School of Electrical and Computer Engineering, Purdue University, West Lafayette, IN, USA. His research interests are in the areas of sensors, the Internet of Things, signal processing, and machine learning.
\end{IEEEbiography}

\begin{IEEEbiography}
[{\includegraphics[width=1in,height=1.25in,clip,keepaspectratio]{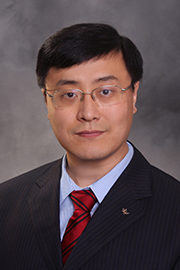}}]
{Lu Su}
(M'15) is an associate professor in the School of Electrical and Computer Engineering at Purdue University. His research interests are in the general areas of Internet of Things and Cyber-Physical Systems, with a current focus on wireless, mobile, and crowd sensing systems. He received Ph.D. in Computer Science, and M.S. in Statistics, both from the University of Illinois at Urbana-Champaign, in 2013 and 2012, respectively. He has also worked at IBM T. J. Watson Research Center and National Center for Supercomputing Applications. He has published more than 100 papers in referred journals and conferences, and serves as an associate editor of ACM Transactions on Sensor Networks. He is the recipient of NSF CAREER Award, University at Buffalo Young Investigator Award, ICCPS’17 best paper award, and the ICDCS’17 best student paper award. He is a member of ACM and IEEE.
\end{IEEEbiography}

\begin{IEEEbiography}
[{\includegraphics[width=1in,height=1.25in,clip,keepaspectratio]{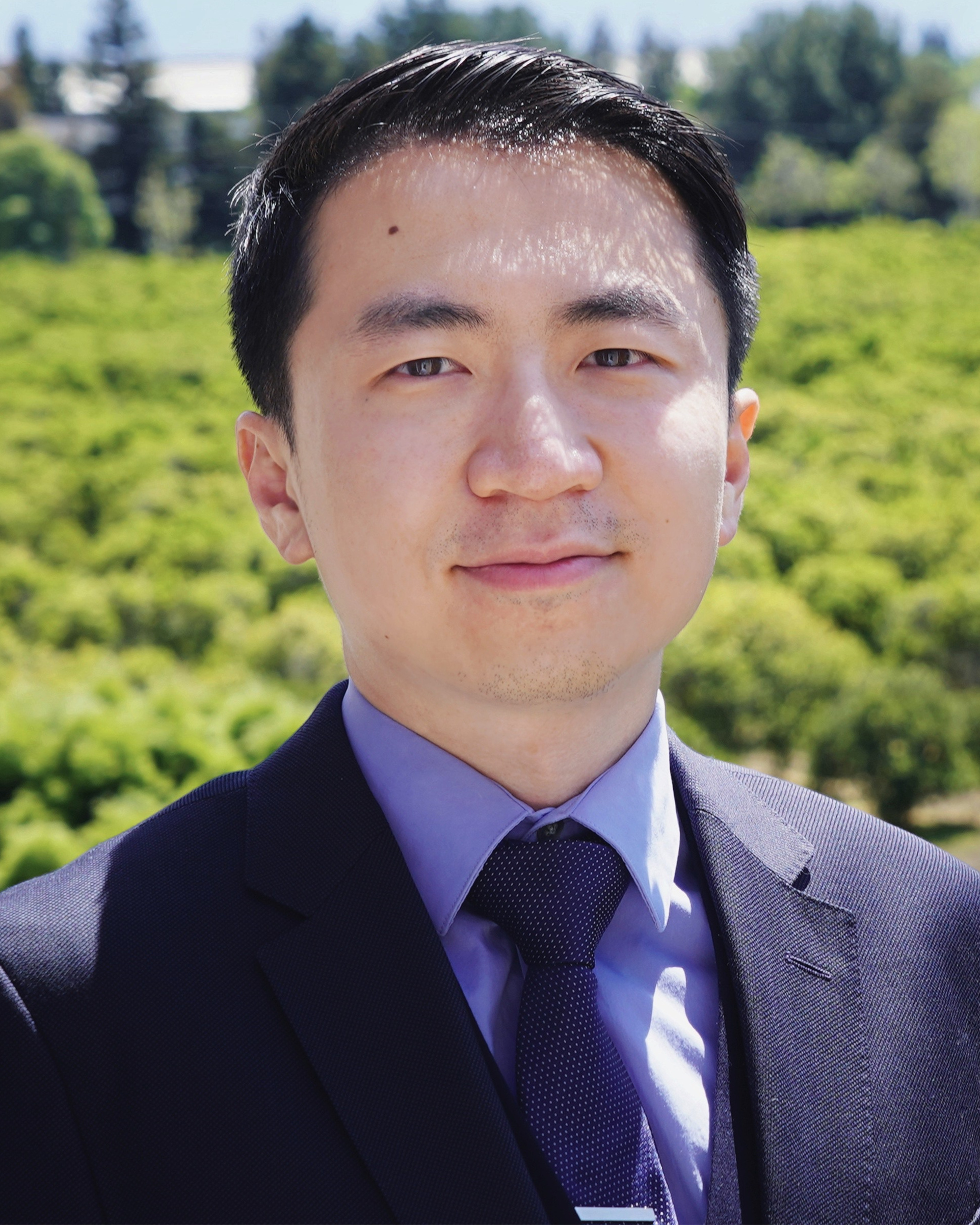}}]
{Ziran Wang}
(S'16-M'19) received the Ph.D. degree from the University of California, Riverside in 2019. He is an Assistant Professor in the College of Engineering at Purdue University, and was a Principal Researcher at Toyota Motor North America. He serves as Founding Chair of IEEE Technical Committee on Internet of Things in Intelligent Transportation Systems, and Associate Editor of four academic journals, including IEEE Internet of Things Journal and IEEE Transactions on Intelligent Vehicles. His research focuses on automated driving, human-autonomy teaming, and digital twin.
\end{IEEEbiography}

\end{document}